\documentclass[conference]{IEEEtran}
\IEEEoverridecommandlockouts
\usepackage{cite}
\usepackage{amsmath,amssymb,amsfonts}
\usepackage{algorithmic}
\usepackage{graphicx}
\usepackage{textcomp}
\usepackage{xcolor}
\usepackage{graphicx}
\usepackage{subfig}   % IEEE-recommended over subfigure
\usepackage{subcaption}
\usepackage{booktabs}
\usepackage{tabularx}
\usepackage{comment}
\usepackage{hyperref}
\newcommand{\linebreakand}{%
  \end{@IEEEauthorhalign}
  \hfill\mbox{}\par
  \mbox{}\hfill\begin{@IEEEauthorhalign}
}

\def\BibTeX{{\rm B\kern-.05em{\sc i\kern-.025em b}\kern-.08em
    T\kern-.1667em\lower.7ex\hbox{E}\kern-.125emX}}
\begin{document}

\title{An Analysis of Active Learning Algorithms using Real-World Crowd-sourced Text Annotations\\

\thanks{This research was supported in part by the National Science Foundation under Grant Number: 2143424 (NSF CAREER Award)}

}

%\author{\IEEEauthorblockN{Anonymous Submission}}

% \author{

% \IEEEauthorblockN{Varun Totakura}
% \IEEEauthorblockA{
% \textit{Department of Computer Science} \\
% \textit{Florida State University} \\
% % Tallahassee, FL, USA \\
% % varun.totakura@gmail.com
% }
% \and

% \IEEEauthorblockN{Ankita Singh}
% \IEEEauthorblockA{
% \textit{Department of Computer Science} \\
% \textit{Florida State University} \\
% % Tallahassee, FL, USA \\
% % ankita@fsu.edu
% }

% \linebreakand

% \IEEEauthorblockN{Yushun Dong}
% \IEEEauthorblockA{
% \textit{Department of Computer Science} \\
% \textit{Florida State University} \\
% % Tallahassee, FL, USA \\
% % yushun@fsu.edu
% }
% \and

% \IEEEauthorblockN{Shayok Chakraborty}
% \IEEEauthorblockA{
% \textit{Department of Computer Science} \\
% \textit{Florida State University} \\
% % Tallahassee, FL, USA \\
% % shayok@fsu.edu
% }

% }

\author{
\IEEEauthorblockN{Varun Totakura, Ankita Singh, Yushun Dong, Shayok Chakraborty}

\IEEEauthorblockA{
\textit{Department of Computer Science} \\
\textit{Florida State University} \\
Tallahassee, FL, USA \\
}
}

\maketitle

\begin{abstract}
Active learning algorithms automatically identify the most informative samples from large amounts of unlabeled data and tremendously reduce human annotation effort in inducing a machine learning model. In a conventional active learning setup, the labeling oracles are assumed to be infallible, that is, they always provide correct answers (in terms of class labels) to the queried unlabeled instances, which cannot be guaranteed in real-world applications. To this end, a body of research has focused on the development of active learning algorithms in the presence of imperfect / noisy oracles. Existing research on active learning with noisy oracles typically simulate the oracles using machine learning models; however, real-world situations are much more challenging, and using ML models to simulate the annotation patterns may not appropriately capture the nuances of real-world annotation challenges. In this research, we first collect annotations of text samples (from 3 benchmark text classification datasets) from crowd-sourced workers through a crowd-sourcing platform. We then conduct extensive empirical studies of 8 commonly used active learning techniques (in conjunction with deep neural networks) using the obtained annotations. Our analyses sheds light on the performance of these techniques under real-world challenges, where annotators can provide incorrect labels, and can also refuse to provide labels. %To the best of our knowledge, this is the first research effort to study the performance of active learning algorithms with real-world annotations, obtained from crowd-sourced workers. 
We hope this research will provide valuable insights that will be useful for the deployment of deep active learning systems in real-world applications. The obtained annotations can be accessed at \url{https://github.com/varuntotakura/al_rcta/}. 
\end{abstract}


\section{Introduction}

The unprecedented growth of digital data in today's world has expanded the possibilities of  solving a variety of real-world problems using computational learning frameworks. However, annotating data with class labels to induce a reliable deep neural network (DNN) model has remained a fundamental bottleneck. \textit{Active Learning (AL)} algorithms alleviate this challenge by intelligently querying the labels of the most informative samples \cite{Settles_2010}. This drastically reduces the human annotation effort, as only a few unlabeled samples that are selected by the algorithm need to be labeled by the oracles \footnote{we use the terms oracles, labelers, annotators and users interchangeably in this paper}. AL has been extensively studied in text mining \cite{tong_support_2000}, computer vision \cite{Yoo_2019}, bioinformatics \cite{Hatice_2010}, anomaly detection \cite{Pimentel_2020} and a variety of other applications, with promising empirical results.  

Conventional AL algorithms assume that the labeling oracles always provide the correct class labels to the queried unlabeled samples. This assumption may not hold for several real-world applications, where we usually have multiple labelers providing different qualities of annotation. For instance, in a crowdsourcing platform like the Amazon Mechanical Turk (AMT), hundreds of annotators are available who provide varying degrees of noisy annotations. This has motivated the development of AL algorithms which operate in the presence of imperfect / noisy annotators \cite{Huang_2017, Zhang_2015, Yan_2016}. %For instance, one common approach to address this problem is to identify the informative unlabeled samples, together with the optimal labeling oracle for each sample, which has depicted promising empirical performance [XX]. 
However, existing techniques typically simulate the imperfect annotators using classification models, such as random forests, SVMs etc. The models are trained on a subset of the training data and their prediction for a queried unlabeled sample is interpreted as the response of an imperfect oracle for that unlabeled sample \cite{Huang_2017, Chakraborty_2020}. Another approach is to cluster the data into $k$ clusters using the $k$-means algorithm, and to assume that each annotator is an expert on a particular cluster, where their labeling coincides with the ground truth; for samples in other clusters, they commit a fixed percentage of labeling errors \cite{Yan_2011}. While these methods offer a more practical solution to AL than the conventional assumption of noise-free annotators, using machine learning algorithms to model the imperfect annotators may not appropriately capture the nuances of real-world annotation challenges. Human annotation behavior is complex, and it may be a challenge to model it using simple machine learning techniques. Further, AL algorithms are most relevant in the training of data-hungry deep neural networks (DNNs), which have depicted commendable performance in a variety of real-world applications. This necessitates a thorough validation of AL algorithms, in conjunction with deep neural networks (deep active learning), using annotations collected from actual human oracles. In this paper, we attempt to address this relevant problem. Our contributions in this work can be summarized as follows: 

\begin{itemize}

\item We collect crowd-sourced annotations for three benchmark text classification datasets through a crowd-sourcing platform, where annotators can provide incorrect annotations, and can also refuse to provide annotations. The obtained annotations are made publicly available to foster further research on this important topic.

\item We study the performance of $8$ different AL techniques (conventional, as well as those designed to operate with imperfect oracles) with DNNs, using the collected annotations. Our analysis provides valuable insights into the performance of AL algorithms under real-world annotation constraints, which will hopefully be helpful for their deployment in practical applications. 

\end{itemize}

%The rest of the paper is organized as follows: we present an overview of related techniques in Section \ref{sec_related}; our data annotation mechanism is discussed in Section \ref{sec_data}; the results of our experiments are presented in Section \ref{sec_expt}; and we conclude with discussions in Section \ref{sec_conc}. 

\section{Related Work}
\label{sec_related}

We present a brief survey of active learning algorithms in general, followed by a survey of AL in the presence of noisy oracles.

\noindent \textbf{Active Learning}: Active Learning (AL) is a well-researched problem in the machine learning literature \cite{Settles_2010}. With the popularity of data-hungry deep neural networks, researchers have studied the problem of deep active learning (DAL), where the goal is to automatically select the informative samples for manual annotation and simultaneously learn discriminating feature representations using a deep neural network \cite{Ren_2021, Khosla_2023, Nuggehalli_2024, Shafir_2025}. Common DAL techniques include a task agnostic scheme which learns a loss prediction function to predict the loss value of an unlabeled sample and queries samples accordingly \cite{Yoo_2019}, a greedy technique to query a \textit{Coreset} of samples that represent the whole dataset \cite{Coreset_Paper}, a sampling technique based on diverse gradient embeddings (\textit{BADGE}) \cite{Badge_Paper}, a strategy based on temporal output discrepancy that queries samples based on the discrepancy of outputs given by the models at different optimization steps during training \cite{TOD_ICCV_2021} and an AL framework that queries unlabeled samples that can provide the most positive influence on model performance \cite{liu2021influence}. Methods based on adversarial training have particularly shown promising performance for deep AL \cite{Sinha_2019, Zhu_2017, Ducoffe_2018}.

\noindent \textbf{AL with Imperfect Oracles}: A body of research has focused on the development of AL algorithms in the presence of noisy annotators, where the goal is to select informative samples, as well as the best annotators for labeling them \cite{Chen_2022}. A few works have assumed a very simplistic setting, such as a single oracle which can provide incorrect labels and can also abstain from labeling \cite{Yan_2016}, or two labeling oracles, one of which always returns the correct label and the other returns incorrect labels with a fixed probability \cite{Zhang_2015, Donmez_2008}.  

For multiple noisy annotators, a common approach is to use relabeling, where an actively queried sample is relabeled multiple times, and the final label is obtained using majority voting \cite{Donmez_2009, Zheng_2010, Ipeirotis_2014}. Zhao \textit{et al.} \cite{Zhao_2011} proposed an incremental relabeling strategy that identified the unlabeled instances that are most informative to label, and also samples that might benefit from relabeling, because their initial labels may be incorrect. The \textit{ActiveLab} algorithm \cite{Goh_2023} is also based on a similar rationale, and balances between relabeling already labeled samples vs. labeling completely new samples. Another strategy to deal with multiple imperfect annotators is to estimate the single best oracle to annotate every queried sample. Yan \textit{et al.} \cite{Yan_2011, Yan_2012} proposed a probabilistic multi-labeler model to compute the accuracy of each labeler and select the most confident labeler for each queried unlabeled sample. Huang \textit{et al.} \cite{Huang_2017} proposed a cost effective active learning (\textit{CEAL}) framework for active sample selection in the presence of multiple noisy oracles. Chakraborty \cite{Chakraborty_2020} proposed an LP based formulation to identify the uncertain and diverse samples for annotation, and the optimal oracle to annotate each queried sample. 

While these methods have depicted promising empirical results, the imperfect annotators in all these techniques were simulated using machine learning models, which may not be able to appropriately capture the complex nature of human annotation behavior. Yan \textit{et al.} \cite{Yan_2011, Yan_2012} analyzed the performance of AL algorithms on scientific text data annotated by human experts. However, their experimental studies have the following limitations: $(i)$ They only addressed binary classification problems, whereas real-world datasets typically involve multiple classes; studying the AL performance on binary problems may not appropriately reflect the challenges and complexities associated with multiple noisy annotators, who may provide incorrect class labels. $(ii)$ They assumed that an annotator will always provide a label for a queried sample. Our collected annotations do not make this assumption, and annotators can even refuse to provide labels to queried instances, which is a very plausible scenario in real-world applications. $(iii)$ These methods do not use deep learning. In today's deep learning era, a study of AL algorithms with noisy annotators, in conjunction with deep neural networks, is necessary, but lacking. 

In this paper, we conduct a thorough analysis of the performance of different active learning algorithms, in conjunction with deep neural networks, with crowd-sourced annotations, on three multi-class textual datasets, where annotators can provide incorrect annotations and can even refuse to provide annotations. \textit{To the best of our knowledge, this has not been studied in the literature}. We hope the insights gained from this research will enable AL to reach practical applications.

\section{Datasets and Annotations}
\label{sec_data}

\noindent \textbf{Datasets.} We used three benchmark text classification datasets in this research: $(i)$ \textbf{AG News} \cite{AGNews_dataset}, which is a collection of news articles with $4$ classes: \textit{world, sports, business} and \textit{science / technology}; $(ii)$ \textbf{Consumer Complaints} \footnote{https://catalog.data.gov/dataset/consumer-complaint-database}, which contains complaints received about financial products and services and consists of $6$ classes: \textit{debt collection, prepaid card/debit card, mortgage, checking/savings account, student loan} and \textit{vehicle loan/lease}; and $(iii)$ \textbf{Wikipedia Movie Plots}\footnote{https://www.kaggle.com/datasets/jrobischon/wikipedia-movie-plots}, containing summary descriptions of movie plots scraped from Wikipedia, with $4$ classes (movie genres): \textit{drama, comedy, horror} and \textit{action}. 

%Their ground truth labels were removed, and the samples were uploaded to the \textit{Upwork} crowd-sourcing platform (https://www.upwork.com/).

\noindent \textbf{Annotator Recruitment.} We used the \textit{Upwork} crowd-sourcing platform (https://www.upwork.com/) due to its wide global reach and the flexibility it offers in managing project-based tasks. The job was made publicly available to eligible workers worldwide, with the condition that they be comfortable reading and understanding English, since the textual data and the instructions were all written in English. Consent was obtained from each worker to participate in the study. No information about the identity of the workers was recorded.

\noindent \textbf{Annotation Protocol.} We selected $3,000$ samples from each of the three datasets at random. The workers were provided with the list of possible classes for each dataset and were asked to annotate each sample with at most one label category. If they were unsure about the label of a sample and wanted to abstain from labeling, they were asked to enter $0$ as the label. After accepting the job, each worker was given $30$ days' time to submit their annotations. Once a job was submitted by a worker, we manually checked the file for errors, in particular, whether each sample was annotated with one of the valid class labels or $0$. In case of any discrepancies, the worker was contacted to fix the issue. Upon a successful submission, each worker was provided with compensation for his / her efforts. Each set of $3,000$ samples was annotated by different crowd-sourced workers ($10$ for AG News and Consumer Complaints, and $9$ for Wikipedia). The data collection protocol was reviewed and approved by an ethical review committee. 

\noindent \textbf{Analysis.} Table \ref{tab_ann_stat} depicts an analysis of the obtained annotations. We note that the annotation accuracy varies quite significantly across different datasets. For the AG News and Consumer Complaints datasets, it was moderately high ($ \approx 74\%$ and $ \approx78\%$ respectively), whereas for the Wikipedia dataset, it was much lower ($ \approx 56\%$). Table \ref{tab_ann_stat} also demonstrates that the annotators refused to annotate some of the samples; however, this percentage is relatively low. %The table also lists the mean correlation among the different annotators (in terms of the provided labels) for each dataset. We note that the annotator correlation also varies significantly and ranges from moderate (for the AG News and Consumer Complaint datasets) to low (for the Wikipedia dataset). These results demonstrate that annotations obtained in real-world applications can contain significant annotation errors and that annotators can be poorly correlated with one another, all of which need to be considered in the development of AL algorithms. 

\begin{table}[ht]
  \centering
  \footnotesize
  \scriptsize
  \resizebox{0.48\textwidth}{!}{
  \begin{tabular}{ccc}
  \toprule
    \textbf{Dataset} & \textbf{Ann Acc (\%)} & \textbf{Not Annotated (\%)} \\
    \midrule
    AG News & $74.13 \pm 19.53$ & $3.41 \pm 4.27$ \\
    \midrule
    Consumer Complaint & $78.63 \pm 10.06$ & $7.90 \pm 7.22$ \\
    \midrule
    Wikipedia & $56.50 \pm 23.62$ & $3.04 \pm 4.03$ \\
    \bottomrule
  \end{tabular}
  }
\caption{Crowd-sourced annotation statistics for each dataset. The values are averaged across all the crowd-sourced annotators for each dataset.}
\label{tab_ann_stat}
\end{table}

%We also conducted an experiment, where we computed the correlation between the predictions furnished by two trained classification models (SVM and Random Forest) and those obtained from the crowd-sourced workers. The results in Table \ref{tab_ann_stat} demonstrate a poor correlation between the two, particularly for the Wikipedia dataset. This implies that machine learning models (as conventionally used in AL research to simulate noisy annotators) may not be able to appropriately capture human annotation behavior, and using ML classifiers to simulate noisy annotators may not be an apt modeling choice. These results show the necessity of validating AL algorithms with real-world annotations collected from human oracles, which is the main objective of this research. 

%Sample annotations are provided in the Appendix (link is provided in Section IV). 
%\textit{The collected crowd-sourced annotations will be made publicly available upon acceptance of this paper, to promote further research on this topic}. 
We now study the performance of AL algorithms with the obtained annotations.

\begin{figure*}[ht]
    \centering
    \subfloat[AG News]{%
        \label{fig_agnews}
        \includegraphics[trim=1.3in 3.2in 1.7in 3.4in, clip, width=0.31\linewidth]
        {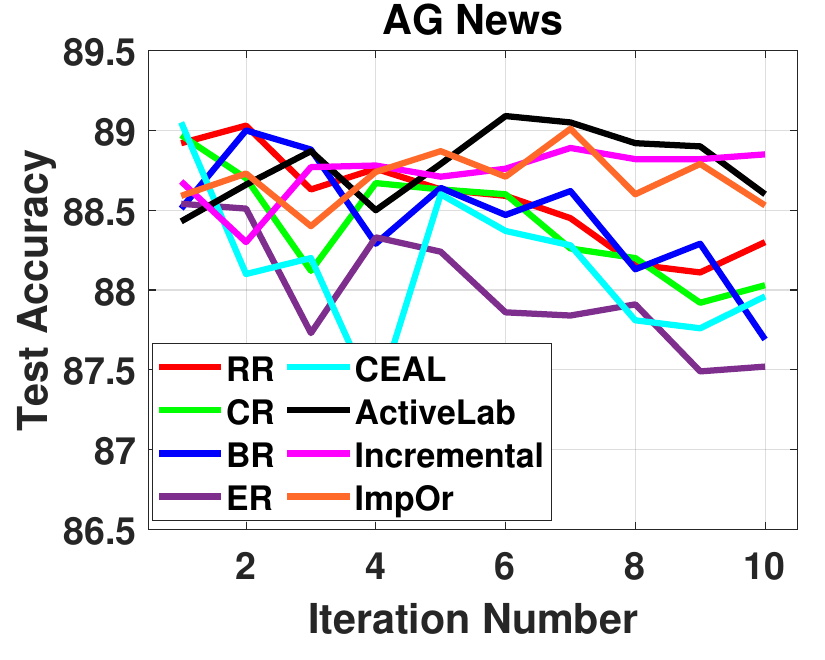}
    }
    \hfill
    \subfloat[Consumer Complaint]{%
        \label{fig_consumer}
        \includegraphics[trim=1.3in 3.2in 1.7in 3.4in, clip, width=0.31\linewidth]
        {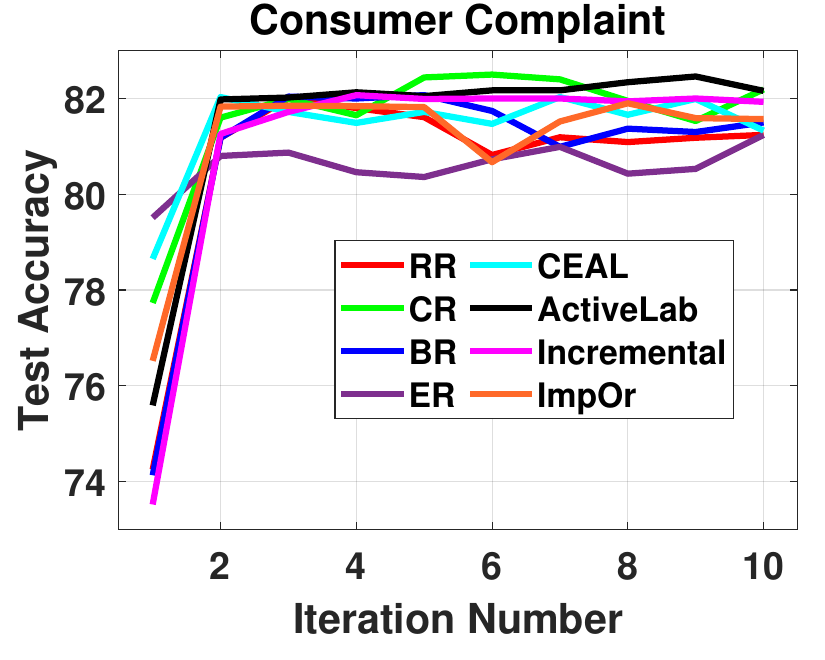}
    }
    \hfill
    \subfloat[Wikipedia Movie Plots]{%
        \label{fig_wiki}
        \includegraphics[trim=1.3in 3.2in 1.7in 3.4in, clip, width=0.31\linewidth]
        {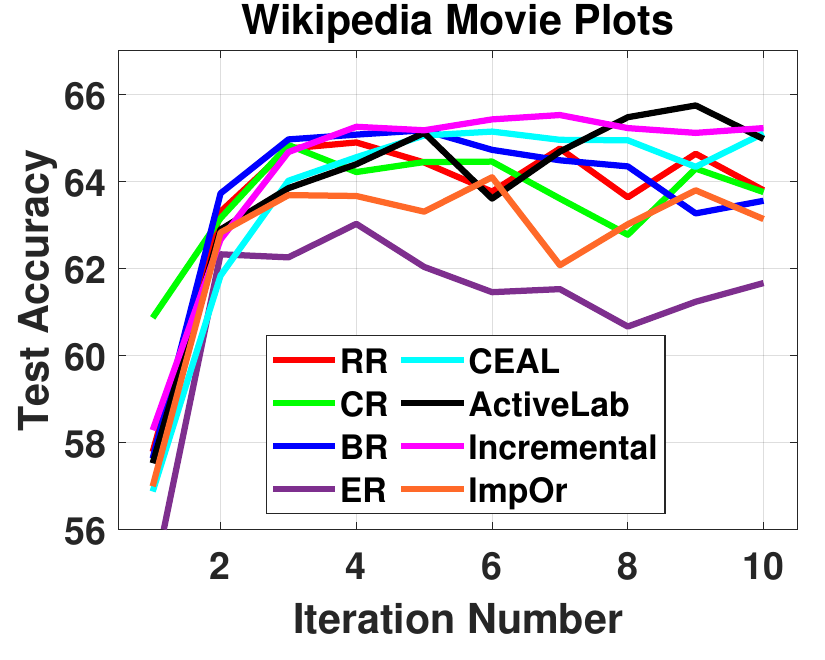}
    }
    \caption{Study of AL performance. Methods that relabel a queried sample multiple times tend to produce better results than methods that label a queried sample using a single best annotator. No single method delivers the best performance consistently. Best viewed in color.}
    \label{fig_main_results}
\end{figure*}

\section{Experiments and Results}
\label{sec_expt}

\noindent \textbf{Experimental Setup.} Each dataset was divided into $3$ parts: $(i)$ an initial training set ($500$ samples) with clean annotations; $(ii)$ an unlabeled set ($2,500$ samples); and $(iii)$ a test set ($10,000$ samples) with clean annotations. A random subset of $2,500$ samples out of the $3,000$ samples whose annotations were obtained from crowd-sourcing, constituted the unlabeled set (the remaining $500$ samples were used to compute the labeling accuracy of each oracle, detailed below). When a (sample-annotator) query was posed by an algorithm, the label provided to the corresponding sample, by the corresponding annotator was retrieved (instead of the ground truth label); the selected sample, together with the label was appended to the training set. If the retrieved label was $0$ (the annotator refused to provide a label to this sample), the sample was simply discarded. After each AL iteration, the model was updated with the augmented training set and tested on the test set. The process was continued iteratively until a stopping condition was satisfied (taken as 10 iterations in this work). The objective was to study the improvement in performance on the test set with increasing sizes of the training set.

Each oracle was assigned an integer between $1$ and $10$ in increasing order of labeling accuracy, which was interpreted as the labeling cost $C_{i}$ of oracle $O_{i}$ for a given dataset, that is, the price to be paid to get one unlabeled sample from the dataset labeled by the particular oracle (more accurate oracles have higher costs). The labeling accuracy of each oracle was computed on the remaining $500$ samples out of the $3,000$ samples (whose annotations were obtained from crowd-sourcing) by comparing the obtained annotations with the ground truth labels. Once a sample was annotated by a given oracle (be it correct or incorrect) the corresponding labeling cost of the oracle was deducted from the available query budget. The query budget $B$ in each AL iteration was set as $100$. All the results were averaged over $3$ runs to rule out the effects of randomness. 

\noindent \textbf{Active Learning Baselines.} We studied the performance of $8$ AL algorithms in this research: (i) \textit{Random}, where a batch of unlabeled samples was queried at random; $(ii)$ \textit{Entropy} \cite{wang_probability_2008}  $(iii)$ \textit{Coreset} \cite{Coreset_Paper} %which selects a set of samples such that a model learned over the selected subset is competitive for the remaining data points; 
$(iv)$ \textit{BADGE} \cite{Badge_Paper} %which samples groups of points with high predictive uncertainty and diversity, that are disparate and high magnitude when represented in a gradient space; 
$(v)$ \textit{CEAL} \cite{Huang_2017} %which iteratively selects a batch of uncertain samples and the optimal annotator to annotate each sample (based on the probability of obtaining the correct label and the cost of the annotator);
$(vi)$ \textit{ActiveLab} \cite{Goh_2023}; $(vii)$ \textit{Incremental} \cite{Zhao_2011}; and $(viii)$ \textit{AL with Imperfect Oracles (ImpOr)} \cite{Chakraborty_2020}. %which selects a batch of uncertain and diverse samples, and the optimal annotator to annotate each sample through an LP formulation. 
The baselines \textit{CEAL, ActiveLab, Incremental} and \textit{ImpOr} are designed to operate with imperfect oracles; they select unlabeled samples, together with corresponding annotator(s) to annotate those samples. Among them, \textit{ActiveLab} and \textit{Incremental} use a relabeling strategy where the same sample can be labeled multiple times by different annotators, while \textit{CEAL} and \textit{ImpOr} select the single best optimal annotator for each queried sample. The other baselines (\textit{Random, Entropy, Badge} and \textit{Coreset}) merely select a batch of unlabeled samples; for these baselines, an oracle was selected at random to annotate the selected samples. They are thus denoted by Random-Random (\textit{RR}), Entropy-Random (\textit{ER}), Badge-Random (\textit{BR}) and Coreset-Random (\textit{CR}) respectively. These baselines were selected to capture the state-of-the-art methods in conventional AL, as well as the two most commonly used strategies in AL with noisy oracles: relabeling and optimal oracle selection. We used the BERT model \cite{BERT_paper} as the backbone DNN architecture in our experiments (for all the methods, for consistency) due to its popularity in text classification applications. 

\noindent \textbf{Implementation Details}: The BERT model was trained using the conventional cross entropy loss; we used a learning rate of $1e^{-5}$ (with cosine annealing), momentum of $0.9$, weight decay of $5e^{-7}$, and the \textit{Adam} optimizer. The experiments were performed on a workstation with 64 GB RAM and two NVIDIA Quadro RTX 5,000 GPUs with 16 GB memory each. The implementations were performed using PyTorch; the Python packages used include Hugging Face Transformers for BERT/Bert Tokenizer, Scikit-Learn, NumPy, Pandas, and Scipy.

\begin{comment}
\begin{table*}[htbp]
\centering
\caption{P-value matrix for AG News dataset.}
\scriptsize
\renewcommand{\arraystretch}{1.3}
\begin{tabular}{|c|c|c|c|c|c|c|c|c|}
\hline
\textbf{} &
\textbf{RR} &
\textbf{ER} &
\textbf{CR} &
\textbf{BR} &
\textbf{CEAL} &
\textbf{ActiveLab} &
\textbf{Incremental} &
\textbf{ImpOr} \\
\hline
\textbf{RR} & -- & -- & -- & -- & -- & -- & -- & -- \\
\hline
\textbf{ER} & $\bold{0.00}$ & -- & -- & -- & -- & -- & -- & -- \\
\hline
\textbf{CR} & $\bold{0.02}$ & $\bold{0.00}$ & -- & -- & -- & -- & -- & -- \\
\hline
\textbf{BR} & $0.14$ & $\bold{0.02}$ & $0.37$ & -- & -- & -- & -- & -- \\
\hline
\textbf{CEAL} & $\bold{0.01}$ & $0.19$ & $\bold{0.04}$ & $\bold{0.04}$ & -- & -- & -- & -- \\
\hline
\textbf{ActiveLab} & $0.08$ & $\bold{0.00}$ & $\bold{0.02}$ & $\bold{0.02}$ & $\bold{0.00}$ & -- & -- & -- \\
\hline
\textbf{Incremental} & $0.11$ & $\bold{0.00}$ & $\bold{0.03}$ & $\bold{0.05}$ & $\bold{0.00}$ & $0.29$ & -- & -- \\
\hline
\textbf{ImpOr} & $0.13$ & $\bold{0.00}$ & $\bold{0.02}$ & $\bold{0.04}$ & $\bold{0.00}$ & $0.15$ & $0.30$ & -- \\
\hline
\end{tabular}
\label{tab_pvalue_ag}
\end{table*}
\end{comment}

\begin{table}[ht]
\centering
\scriptsize
\resizebox{0.48\textwidth}{!}{
\begin{tabular}{lcccccccc}
\toprule
\textbf{} & \textbf{RR} & \textbf{ER} & \textbf{CR} & \textbf{BR} & \textbf{CEAL} & \textbf{ActiveLab} & \textbf{Incremental} & \textbf{ImpOr}\\
%\midrule
\textbf{RR}        &         &         &         &        &         &         &        \\
\\
\textbf{ER}       & $\bold{0.00}$  &         &         &        &         &         &        \\
\\
\textbf{CR}       & $\bold{0.02}$  &  $\bold{0.00}$    &         &        &         &         &        \\
\\
\textbf{BR}         & $0.14$  & $\bold{0.02}$ & $0.37$  &         &        &         &         &        \\
\\
\textbf{CEAL}          & $\bold{0.01}$  & $0.19$  & $\bold{0.04}$  & $\bold{0.04}$  &        &         &         &        \\
\\
\textbf{ActiveLab}     & $0.08$  & $\bold{0.00}$  & $\bold{0.02}$  & $\bold{0.02}$  & $\bold{0.00}$ &         &         &        \\
\\
\textbf{Incremental}   & $0.11$  & $\bold{0.00}$  & $\bold{0.03}$  & $\bold{0.05}$  & $\bold{0.00}$ & $0.29$  &         &        \\
\\
\textbf{ImpOr}            & $0.13$  & $\bold{0.00}$  & $\bold{0.02}$  & $\bold{0.04}$  & $\bold{0.00}$ & $0.15$  & $0.30$  &        \\
\bottomrule
\end{tabular}
}
\caption{P-value matrix: AG News.}
\label{tab_pvalue_ag}
\end{table}

\begin{comment}
\begin{table*}[htbp]
\centering
\caption{P-value matrix for Consumer Complaint dataset.}
\scriptsize
\renewcommand{\arraystretch}{1.3}
\begin{tabular}{|c|c|c|c|c|c|c|c|c|}
\hline
\textbf{} &
\textbf{RR} &
\textbf{ER} &
\textbf{CR} &
\textbf{BR} &
\textbf{CEAL} &
\textbf{ActiveLab} &
\textbf{Incremental} &
\textbf{ImpOr} \\
\hline
\textbf{RR} & -- & -- & -- & -- & -- & -- & -- & -- \\
\hline
\textbf{ER} & $0.47$ & -- & -- & -- & -- & -- & -- & -- \\
\hline
\textbf{CR} & $\bold{0.01}$ & $\bold{0.01}$ & -- & -- & -- & -- & -- & -- \\
\hline
\textbf{BR} & $\bold{0.04}$ & $0.36$ & $\bold{0.02}$ & -- & -- & -- & -- & -- \\
\hline
\textbf{CEAL} & $\bold{0.05}$ & $\bold{0.02}$ & $0.18$ & $0.12$ & -- & -- & -- & -- \\
\hline
\textbf{ActiveLab} & $\bold{0.00}$ & $0.06$ & $0.36$ & $\bold{0.00}$ & $0.39$ & -- & -- & -- \\
\hline
\textbf{Incremental} & $\bold{0.02}$ & $0.27$ & $0.11$ & $0.10$ & $0.25$ & $\bold{0.02}$ & -- & -- \\
\hline
\textbf{ImpOr} & $\bold{0.03}$ & $0.12$ & $\bold{0.02}$ & $0.17$ & $0.11$ & $\bold{0.04}$ & $0.42$ & -- \\
\hline
\end{tabular}
\label{tab_pvalue_cc}
\end{table*}
\end{comment}

\begin{table}[ht]
\centering
\scriptsize
\resizebox{0.48\textwidth}{!}{
\begin{tabular}{lcccccccc}
\toprule
\textbf{} & \textbf{RR} & \textbf{ER} & \textbf{CR} & \textbf{BR} & \textbf{CEAL} & \textbf{ActiveLab} & \textbf{Incremental} & \textbf{ImpOr}\\
%\midrule
\textbf{RR}        &         &         &         &        &         &         &        \\
\\
\textbf{ER}       & $0.47$  &         &         &        &         &         &        \\
\\
\textbf{CR}       & $\bold{0.01}$  & $\bold{0.01}$  &         &        &         &         &        \\
\\
\textbf{BR}         & $\bold{0.04}$  & $0.36$  & $\bold{0.02}$  &         &        &         &         &        \\
\\
\textbf{CEAL}          & $\bold{0.05}$  & $\bold{0.02}$  & $0.18$  & $0.12$  &        &         &         &        \\
\\
\textbf{ActiveLab}     & $\bold{0.00}$  & $0.06$  & $0.36$  & $\bold{0.00}$  & $0.39$ &         &         &        \\
\\
\textbf{Incremental}   & $\bold{0.02}$  & $0.27$  & $0.11$  & $0.10$  & $0.25$ & $\bold{0.02}$  &         &        \\
\\
\textbf{ImpOr}            & $\bold{0.03}$  & $0.12$  & $\bold{0.02}$  & $0.17$  & $0.11$ & $\bold{0.04}$  & $0.42$  &        \\
\bottomrule
\end{tabular}
}
\caption{P-value matrix: Consumer Complaint.}
\label{tab_pvalue_cc}
\end{table}

\begin{comment}
\begin{table*}[htbp]
\centering
\caption{P-value matrix for Wikipedia Movie Plots dataset.}
\scriptsize
\renewcommand{\arraystretch}{1.3}
\begin{tabular}{|c|c|c|c|c|c|c|c|c|}
\hline
\textbf{} &
\textbf{RR} &
\textbf{ER} &
\textbf{CR} &
\textbf{BR} &
\textbf{CEAL} &
\textbf{ActiveLab} &
\textbf{Incremental} &
\textbf{ImpOr} \\
\hline
\textbf{RR} & -- & -- & -- & -- & -- & -- & -- & -- \\
\hline
\textbf{ER} & $\bold{0.00}$ & -- & -- & -- & -- & -- & -- & -- \\
\hline
\textbf{CR} & $0.43$ & $\bold{0.00}$ & -- & -- & -- & -- & -- & -- \\
\hline
\textbf{BR} & $0.29$ & $\bold{0.00}$ & $0.45$ & -- & -- & -- & -- & -- \\
\hline
\textbf{CEAL} & $0.37$ & $\bold{0.00}$ & $0.47$ & $0.48$ & -- & -- & -- & -- \\
\hline
\textbf{ActiveLab} & $0.19$ & $\bold{0.00}$ & $0.36$ & $0.36$ & $0.29$ & -- & -- & -- \\
\hline
\textbf{Incremental} & $\bold{0.01}$ & $\bold{0.00}$ & $0.10$ & $\bold{0.03}$ & $\bold{0.00}$ & $\bold{0.04}$ & -- & -- \\
\hline
\textbf{ImpOr} & $\bold{0.00}$ & $\bold{0.00}$ & $\bold{0.01}$ & $\bold{0.00}$ & $\bold{0.01}$ & $\bold{0.00}$ & $\bold{0.00}$ & -- \\
\hline
\end{tabular}
\label{tab_pvalue_wiki}
\end{table*}
\end{comment}

\begin{table}[ht]
\centering
\scriptsize
\resizebox{0.48\textwidth}{!}{
\begin{tabular}{lcccccccc}
\toprule
\textbf{} & \textbf{RR} & \textbf{ER} & \textbf{CR} & \textbf{BR} & \textbf{CEAL} & \textbf{ActiveLab} & \textbf{Incremental} & \textbf{ImpOr}\\
%\midrule
\textbf{RR}        &         &         &         &        &         &         &        \\
\\
\textbf{ER}       & $\bold{0.00}$  &         &         &        &         &         &        \\
\\
\textbf{CR}       & $0.43$  & $\bold{0.00}$  &         &         &        &         &         &        \\
\\
\textbf{BR}         & $0.29$  & $\bold{0.00}$  & $0.45$  &         &        &         &         &        \\
\\
\textbf{CEAL}          & $0.37$  & $\bold{0.00}$  & $0.47$  & $0.48$  &        &         &         &        \\
\\
\textbf{ActiveLab}     & $0.19$  & $\bold{0.00}$  & $0.36$  & $0.36$  & $0.29$ &         &         &        \\
\\
\textbf{Incremental}   & $\bold{0.01}$  & $\bold{0.00}$  & $0.10$  & $\bold{0.03}$  & $\bold{0.00}$ & $\bold{0.04}$  &         &        \\
\\
\textbf{ImpOr}            & $\bold{0.00}$  & $\bold{0.00}$  & $\bold{0.01}$  & $\bold{0.00}$  & $\bold{0.01}$ & $\bold{0.00}$  & $\bold{0.00}$  &        \\
\bottomrule
\end{tabular}
}
\caption{P-value matrix: Wikipedia Movie Plots.}
\label{tab_pvalue_wiki}
\end{table}

\subsection{Study of Active Learning Performance}
\label{subsec_AL_perf}

The active learning performance results are depicted in Figure \ref{fig_main_results}. In each graph, the $x$-axis denotes the AL iteration number and the $y$-axis denotes the accuracy on the test set. We also conducted statistical tests of significance using \textit{paired t-tests} to assess whether the difference in performance between two methods is statistically significant. Tables \ref{tab_pvalue_ag}, \ref{tab_pvalue_cc} and \ref{tab_pvalue_wiki} report the p-values between every pair of methods studied for the three datasets. 

\textbf{AG News}. \textit{ActiveLab}, \textit{Incremental} and \textit{ImpOr} depict the best performance for this dataset. They, in general, outperform the methods that do not consider sample relabeling or optimal oracle selection (\textit{RR, ER, CR} and \textit{BR}); the improvement in performance over \textit{ER, CR} and \textit{BR} is statistically significant $(p < 0.05)$. Among \textit{ActiveLab, Incremental} and \textit{ImpOr}, the difference in performance is not statistically significant $(p > 0.05)$. \textit{CEAL}, even though considers optimal oracle selection, does not perform well. 

\textbf{Consumer Complaint}. \textit{ActiveLab} outperforms \textit{Incremental} and \textit{ImpOr} at a significant level $(p < 0.05)$. It also outperforms \textit{CEAL} in the later AL iterations, although the improvement is not statistically significant. All the methods outperform \textit{RR} at a significant level; they also outperform \textit{BR}. 

\textbf{Wikipedia}. \textit{Incremental} depicts the best performance for this dataset; its improvement in performance is statistically significant for all the methods except \textit{CR}. The methods that consider optimal oracle / sample relabeling (\textit{ActiveLab, Incremental, CEAL}) tend to outperform the other methods, where the annotators are selected at random. \textit{ImpOr}, however, depicts the worst performance, even though it considers optimal annotator selection. %This is probably because 

\textbf{Insights.} We derive the following insights from our results: %\yd{It would make these observations read great if we can imply some structure for these observations to be introduced below, e.g., from general to specific or from commonly seen/widely acknowledged to interesting/less mentioned or studied ones.}

\begin{itemize}

\item No single method delivers the best performance consistently across all the datasets, at a statistically significant level. 

\item AL algorithms designed to operate with imperfect annotators tend to outperform methods that assume the oracles to be infallible and focus on querying the informative samples only. 

\item Methods that relabel a queried sample multiple times using different annotators (\textit{ActiveLab, Incremental}) tend to produce better results than methods that label a queried sample using only a single best annotator (\textit{CEAL, ImpOr}). Thus, in a real-world setup with noisy annotators, it is probably a better strategy to annotate a given sample by multiple oracles and collate the annotations to derive the final label, rather than attempting to find the single optimal annotator for the sample. 

\item From Figure \ref{fig_main_results}, we note that the AL curves for Consumer Complaint and Wikipedia depict an increasing trend, whereas those for AG News depict more or less a constant pattern. To study this further, we conducted experiments using the \textit{ActiveLab} method with $50$ and $100$ training samples to start with (instead of $500$, as in the original experiment). The results, shown in Figure \ref{fig_initial_train_results}, depict an increasing trend in accuracy. This corroborates that annotating samples beyond a certain point may not increase the accuracy (and may even adversely affect the accuracy). Thus, automatically deriving an appropriate stopping condition (when to stop annotating samples) is an important consideration in AL. While there has been some research in this area \cite{AL_Stopping_1, AL_Stopping_2, AL_Stopping_3}, it has not been explored in the context of noisy annotators. Conventionally, samples are annotated until the query budget is exhausted, which may not furnish the optimal performance always. 

\begin{figure}[ht]
	\centering
          \label{fig_agnews_GPT}
          \includegraphics[trim = 1.3in 3.2in 1.2in 3.4in,clip,width=.3\textwidth]{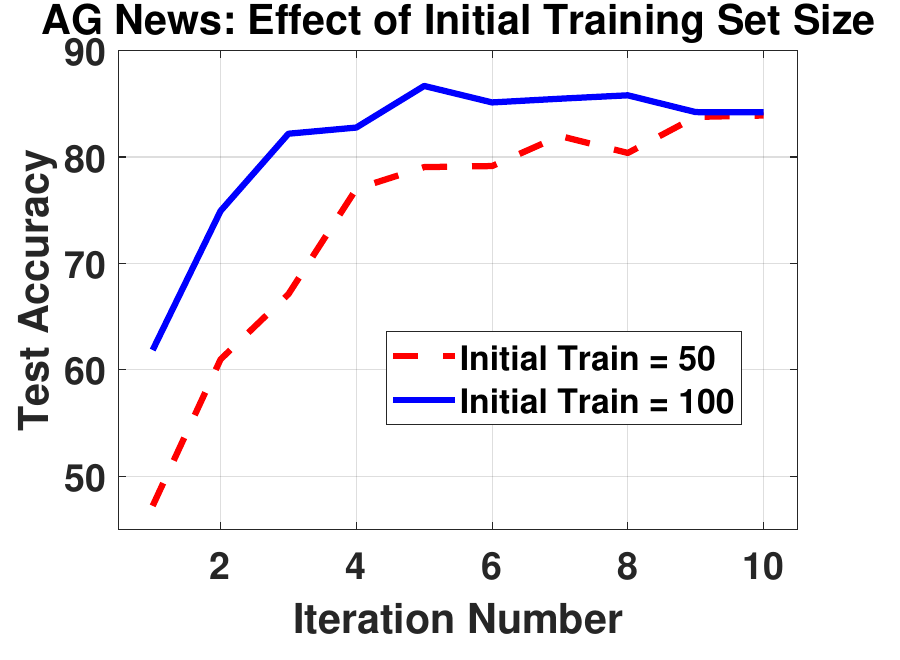}     
          \caption{Effect of initial training set size: \textit{ActiveLab} on AG News dataset. Best viewed in color.}
        \label{fig_initial_train_results}     
\end{figure} 

\begin{comment}
\begin{table}[htbp]
\centering
\caption{Number of samples (out of 3,000) that were incorrectly and correctly annotated by all annotators for each dataset.}
\scriptsize
\renewcommand{\arraystretch}{1.3}
\begin{tabular}{|c|c|c|}
\hline
\textbf{Dataset} & \textbf{All Incorrect} & \textbf{All Correct} \\
\hline
AG News & 89 & 272 \\
\hline
Consumer Complaint & 36 & 671 \\
\hline
Wikipedia & 16 & 19 \\
\hline
\end{tabular}
\label{tab_ann_acc}
\end{table}
\end{comment}

\begin{table}[ht]
  \centering
  \footnotesize
  \resizebox{0.45\textwidth}{!}{
  \begin{tabular}{lcc}
  \toprule
    \textbf{Dataset} & \textbf{All Incorrect} & \textbf{All Correct}  \\
    \midrule
    AG News & $89$ & $272$ \\
    \midrule
    Consumer Complaint & $36$ & $671$ \\
    \midrule
    Wikipedia & $16$ & $19$ \\
    \bottomrule
  \end{tabular}
  }
\caption{Number of samples (out of $3,000$) that were incorrectly and correctly annotated by all the annotators, for each dataset.}
\label{tab_ann_acc}
\end{table}

\item Our analysis further revealed that certain samples in each dataset were labeled incorrectly by all the annotators consistently. Table \ref{tab_ann_acc} reports the number of samples in each dataset that were annotated incorrectly and correctly by all the annotators. An example from the AG News dataset, with all incorrect annotations, is shown below: %\yd{dangling word -- we'd better avoid them throughout the paper. Also, there seems to be no transition from the observations above to the discussion starting here till the end of this section. Do we have a way to have the audience better prepared for it/them?}

\textit{``Currency traders, investors and strategists are more bearish on the dollar than at any time in the past 18 months,
a Bloomberg News survey indicates.''}

This sample actually belongs to class \textit{``World''}, but was labeled as \textit{``Business''} by $9$ annotators and \textit{``Sports''} by one annotator. This is perhaps understandable, as the class definitions of \textit{``World''} and \textit{``Business''} may share some overlap, and it may be difficult to distinguish samples of one class from the other. Such samples can be extremely difficult to handle, as they are labeled not only incorrectly, but also consistently by most of the annotators; thus, even AL techniques that relabel a sample multiple times with different annotators will fail to derive the correct label for such instances. A possible solution may be to allow annotators to specify a \textit{second-best} preference along with their top preference, when they annotate a sample, and develop an AL technique which considers \textit{ordered class annotations}. There is very little research in this area \cite{Xue_2019}, and it is an interesting direction for future AL research. 

\end{itemize}

\begin{table*}[ht]
  \centering
  \footnotesize
  \scriptsize  % Reduced font size further than \footnotesize
  \renewcommand{\arraystretch}{1.7}  % Increases vertical spacing between rows
  \resizebox{0.8\textwidth}{!}{ %
    \begin{tabular}{c|c|c|c}
  \toprule
  \textbf{Method} & \textbf{AG News} & \textbf{Consumer Complaint} & \textbf{Wikipedia} \\
  \midrule
  \textbf{RR}           & $18.68 \pm 2.81$ $(13.26 \pm 3.42)$ & $18.8 \pm 2.16$ $(13.53 \pm 2.74)$ & $20.06 \pm 1.74$ $(11.43 \pm 2.12)$ \\
  \textbf{ER}           & $18.20 \pm 3.32$ $(12.10 \pm 3.31)$ & $19.20 \pm 2.09$ $(12.90 \pm 2.07)$ & $20.30 \pm 2.62$ $(10.12 \pm 1.78)$\\
  \textbf{CR}           & $18.26 \pm 2.04$ $(13.66 \pm 2.28)$ & $18.26 \pm 2.04$ $(12.77 \pm 3.28)$ & $20.46 \pm 1.92$ $(11.1 \pm 2.24)$ \\
  \textbf{BR}           & $18.6 \pm 2.15$ $(13.23 \pm 2.51)$  & $19.26 \pm 2.37$ $(14.03 \pm 2.09)$ & $20.5 \pm 1.89$ $(11.66 \pm 2.49)$ \\
  \textbf{CEAL}         & $16.36 \pm 2.72$ $(7.7 \pm 2.45)$  & $10.13 \pm 0.44$ $(8.23 \pm 1.28)$  & $17.0 \pm 0.0$ $(13.97 \pm 1.85)$ \\
  \textbf{ActiveLab}    & $18.43 \pm 0.76$ $(6.56 \pm 3.48)$  & $19.02 \pm 0.08$ $(8.03 \pm 3.92)$     & $23.33 \pm 2.25$ $(10.16 \pm 2.54)$ \\
  \textbf{Incremental}  & $4.33 \pm 0.53$ $(3.33 \pm 0.53)$   & $4.23 \pm 0.49$ $(3.06 \pm 0.68)$    & $4.07 \pm 0.25$ $(3.03 \pm 0.73)$ \\
  \textbf{ImpOr}        & $10.08 \pm 0.46$ $(2.86 \pm 1.82)$     & $14.23 \pm 1.09$ $(4.5 \pm 1.65)$     & $27.63 \pm 3.76$ $(13.03 \pm 3.37)$ \\
  \bottomrule
\end{tabular}
}
\caption{Study of labeling accuracy. Average number of samples queried (average number of samples correctly annotated) in each AL iteration for each method studied.}
\label{tab_labeling_acc}
\end{table*}

\begin{comment}
\begin{table*}[htbp]
\centering
\caption{Computation time analysis. Average time required (in seconds) to query a batch of unlabeled samples (one AL iteration) for each method studied.}
\renewcommand{\arraystretch}{1.3}
\footnotesize
\resizebox{\textwidth}{!}{
\begin{tabular}{|c|c|c|c|c|c|c|c|c|}
\hline
\textbf{Dataset} 
& \textbf{RR} 
& \textbf{ER} 
& \textbf{CR} 
& \textbf{BR} 
& \textbf{CEAL} 
& \textbf{ActiveLab} 
& \textbf{Incremental} 
& \textbf{ImpOr} \\
\hline
AG News 
& $4.23 \pm 1.38$ 
& $12.79 \pm 3.21$ 
& $35.62 \pm 4.91$ 
& $25.53 \pm 3.09$ 
& $1465.93 \pm 83.12$ 
& $106.59 \pm 41.39$ 
& $30.07 \pm 9.26$ 
& $1338.19 \pm 45.26$ \\
\hline
Consumer Complaint 
& $5.62 \pm 3.02$ 
& $17.57 \pm 1.09$ 
& $35.66 \pm 5.23$ 
& $28.08 \pm 2.51$ 
& $1304.86 \pm 298.05$ 
& $92.34 \pm 31.20$ 
& $32.86 \pm 8.67$ 
& $1390.40 \pm 169.31$ \\
\hline
Wikipedia 
& $2.27 \pm 1.82$ 
& $18.69 \pm 1.27$ 
& $37.06 \pm 4.45$ 
& $23.54 \pm 3.01$ 
& $809.49 \pm 300.97$ 
& $80.72 \pm 24.54$ 
& $27.23 \pm 9.49$ 
& $2248.11 \pm 1891.68$ \\
\hline
\end{tabular}
}
\label{tab_complexity}
\end{table*}
\end{comment}

\begin{table*}[ht]
  \centering
  \footnotesize
  \resizebox{\textwidth}{!}{
  \begin{tabular}{ccccccccc}
  \toprule
    \textbf{Dataset} & \textbf{RR} & \textbf{ER} & \textbf{CR} & \textbf{BR} & \textbf{CEAL} & \textbf{ActiveLab} & \textbf{Incremental} & \textbf{ImpOr} \\
    \midrule
    AG News & $4.23 \pm 1.38$ & $12.79 \pm 3.21$ & $35.62 \pm 4.91$ & $25.53 \pm 3.09$ & $1465.93 \pm 83.12$ & $106.59 \pm 41.39$ & $30.07 \pm 9.26$ & $1338.19 \pm 45.26$ \\
    \midrule
    Consumer Complaint & $5.62 \pm 3.02$ & $17.57 \pm 1.09$ & $35.66 \pm 5.23$ & $28.08 \pm 2.51$ & $1304.86 \pm 298.05$ & $92.34 \pm 31.20$ & $32.86 \pm 8.67$ & $1390.40 \pm 169.31$ \\
    \midrule
    Wikipedia & $2.27 \pm 1.82$ & $18.69 \pm 1.27$ & $37.06 \pm 4.45$ & $23.54 \pm 3.01$ & $809.49 \pm 300.97$ & $80.72 \pm 24.54$ & $27.23 \pm 9.49$ & $2248.11 \pm 1891.68$ \\
    \bottomrule
  \end{tabular}
  }
\caption{Computation time analysis. Average time required (in seconds) to query a batch of unlabeled samples (one AL iteration) for each method studied.}
\label{tab_complexity}
\end{table*}

\subsection{Study of Labeling Accuracy}

The goal of this experiment was to study the labeling accuracy of each of the AL methods. Table \ref{tab_labeling_acc} shows the average number of samples queried in each AL iteration by each method, together with the average number of samples that were annotated correctly (in parentheses). 

\textbf{Insights.} We derive the following insights from our results:

\begin{itemize}

\item Methods that use a random oracle (\textit{RR, ER, BR, CR}) query $\approx20$ samples in each iteration. This is expected, as the query budget is $100$ and the oracles have a cost of $1$ to $10$; thus, the expected cost of annotation for each sample is $5$, which amounts to $20$ queries in each AL iteration. 

\item \textit{Incremental} queries each sample multiple times from different annotators; since a price needs to be paid for each query, it can query only a few samples in each AL iteration. However, due to the multiple annotations, it has a very high success rate. The other methods query more samples, but their success rate is much lower; thus, they end up polluting the training set with incorrectly labeled samples, which can potentially degrade their performance. For instance, for Wikipedia, \textit{ImpOr} queries $\approx27$ samples in each run, out of which only $\approx13$ are correctly annotated; thus, it introduces about $14$ incorrectly labeled samples in each iteration, which accounts for its poor performance (Figure \ref{fig_wiki}). \textit{Incremental} queries only $4$ samples, out of which $3$ are correctly annotated and depicts much better performance. Thus, querying a small, but correctly annotated batch may sometimes outperform querying larger batches with noisier annotations, particularly for datasets like Wikipedia, where the annotation accuracy is low (Table \ref{tab_ann_stat}).

%\item In some cases, selecting an annotator at random may result in a better annotation accuracy than using sophisticated algorithms to select the annotator for a queried sample. For instance, for Consumer Complaint, \textit{CR} queries $\approx18$ samples in each AL iteration, out of which $\approx12$ are correct annotated (annotation accuracy   $\approx 66.67\%$), while \textit{ImpOr} (which attempts to find the best annotator for a given sample) queries $\approx14$ samples out of which only $\approx4$ are correctly annotated (annotation accuracy $\approx 28.57\%$), explaining \textit{CR}'s better performance than \textit{ImpOr} in Figure \ref{fig_consumer}. 

\end{itemize}

\begin{figure*}[ht]
    \centering
       \subfloat[Budget 50]{%
        \label{fig_agnews_budget_50}
        \includegraphics[
            trim=1.3in 3.2in 1.7in 3.4in,
            clip,
            width=0.31\linewidth
        ]{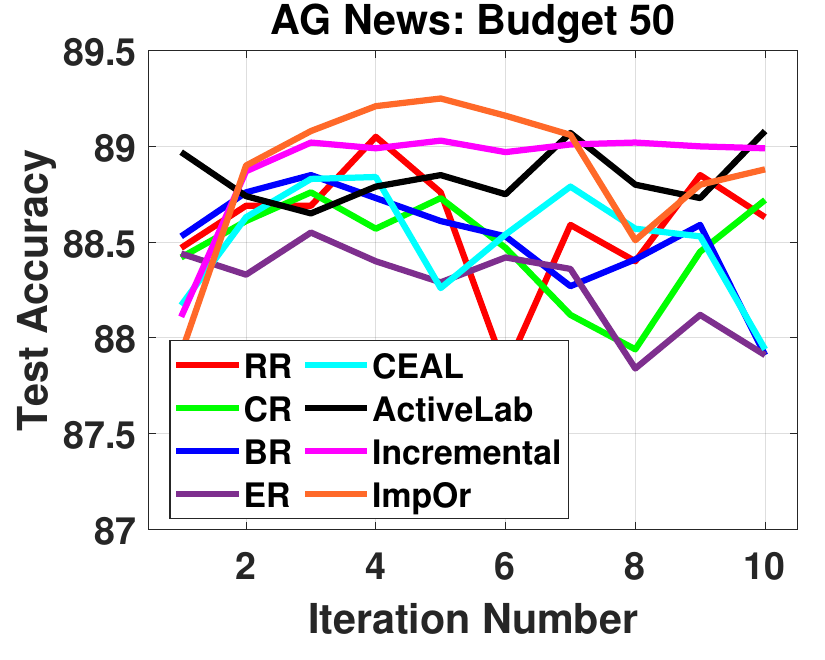}
    }
    \hfill
    \subfloat[Budget 200]{%
        \label{fig_agnews_budget_200}
        \includegraphics[
            trim=1.3in 3.2in 1.7in 3.4in,
            clip,
            width=0.31\linewidth
        ]{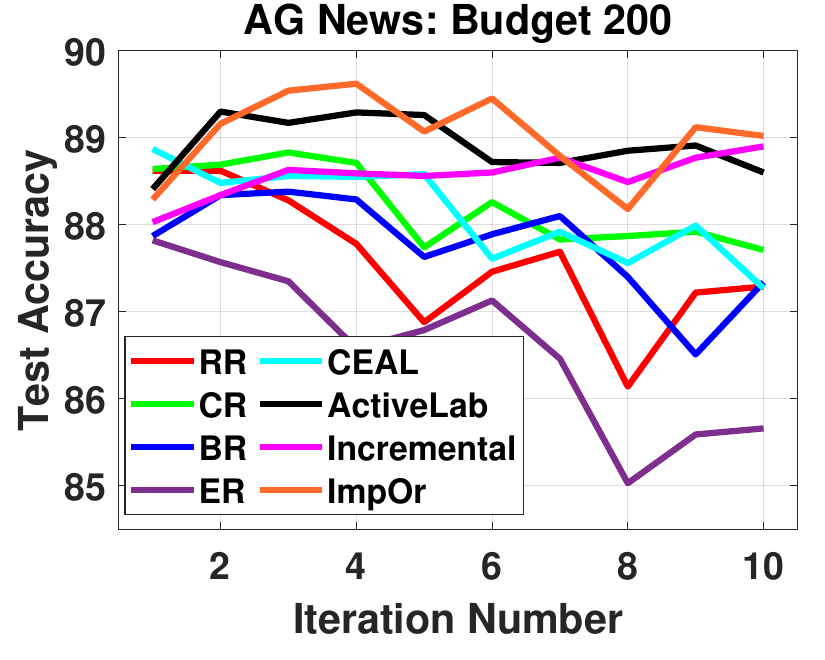}
    }
    \hfill
    \subfloat[Budget 300]{%
        \label{fig_agnews_budget_300}
        \includegraphics[
            trim=1.3in 3.2in 1.7in 3.4in,
            clip,
            width=0.31\linewidth
        ]{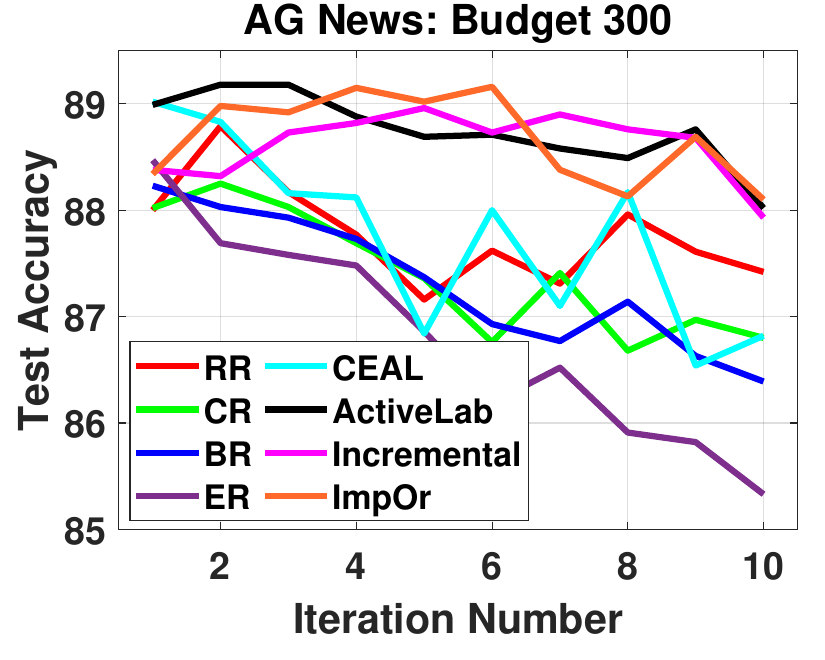}
    }
    \caption{Study of labeling budget on the AG News dataset. \textit{ActiveLab}, \textit{Incremental}, and \textit{ImpOr} outperform the other baselines across all budgets. Please refer to the text for more details. Best viewed in color.}
    \label{fig_budget_results}
\end{figure*}

\subsection{Computation Time Analysis}

Table \ref{tab_complexity} reports the average time required to query a batch of unlabeled samples (one AL iteration) for each method. \textit{RR} has the least computation time, as it performs random selection of samples and annotators and does not involve any computations. \textit{CEAL} evaluates all possible (sample-annotator) pairs and selects the optimal pair iteratively, and thus has a high computation time. \textit{ImpOr} solves an LP problem over a large matrix, which increases the computation time. \textit{ActiveLab} and \textit{Incremental} employ an uncertainty based active sampling criterion together with multiple random oracle assignments and thus have much lower computation time. \textit{ER} involves an entropy computation for each unlabeled sample and also has a low computation time. \textit{BADGE} performs a gradient embedding and a sampling computation, followed by \textit{k-means++} for sample selection; \textit{Coreset} solves a mixed integer programming problem for sample selection. These operations also require moderate computation time, as reflected in Table \ref{tab_complexity}. %for \textit{BR} and \textit{CR}. 

\begin{comment}
\begin{table}[ht]
\centering
\caption{P-value matrix: AG News: Budget $50$.}
\scriptsize
\renewcommand{\arraystretch}{1.3}
\resizebox{0.48\textwidth}{!}{
\begin{tabular}{|c|c|c|c|c|c|c|c|c|}
\hline
\textbf{} 
& \textbf{RR} 
& \textbf{ER} 
& \textbf{CR} 
& \textbf{BR} 
& \textbf{CEAL} 
& \textbf{ActiveLab} 
& \textbf{Incremental} 
& \textbf{ImpOr} \\
\hline
\textbf{RR} 
&  &  &  &  &  &  &  &  \\
\hline
\textbf{ER} 
& $\bold{0.01}$ &  &  &  &  &  &  &  \\
\hline
\textbf{CR} 
& $0.16$ & $\bold{0.02}$ &  &  &  &  &  &  \\
\hline
\textbf{BR} 
& $0.27$ & $\bold{0.00}$ & $0.35$ &  &  &  &  &  \\
\hline
\textbf{CEAL} 
& $0.26$ & $\bold{0.01}$ & $0.41$ & $0.46$ &  &  &  &  \\
\hline
\textbf{ActiveLab} 
& $\bold{0.03}$ & $\bold{0.00}$ & $\bold{0.01}$ & $\bold{0.02}$ & $\bold{0.01}$ &  &  &  \\
\hline
\textbf{Incremental} 
& $\bold{0.01}$ & $\bold{0.01}$ & $\bold{0.01}$ & $\bold{0.01}$ & $\bold{0.00}$ & $0.31$ &  &  \\
\hline
\textbf{ImpOr} 
& $\bold{0.04}$ & $\bold{0.01}$ & $\bold{0.01}$ & $\bold{0.02}$ & $\bold{0.01}$ & $0.41$ & $0.37$ &  \\
\hline
\end{tabular}
}
\label{tab_pvalue_ag_50}
\end{table}
\end{comment}

\begin{table}[ht]
\centering
\scriptsize
\resizebox{0.48\textwidth}{!}{
\begin{tabular}{lcccccccc}
\toprule
\textbf{} & \textbf{RR} & \textbf{ER} & \textbf{CR} & \textbf{BR} & \textbf{CEAL} & \textbf{ActiveLab} & \textbf{Incremental} & \textbf{ImpOr}\\
%\midrule
\textbf{RR}        &           &           &          &         &          &         &   \\
\\
\textbf{ER}       & $\bold{0.01}$    &           &          &         &          &         &   \\
\\
\textbf{CR}       & $0.16$    & $\bold{0.02}$    &           &          &         &          &         &   \\
\\
\textbf{BR}         & $0.27$    & $\bold{0.00}$    & $0.35$    &          &         &          &         &   \\
\\
\textbf{CEAL}          & $0.26$    & $\bold{0.01}$    & $0.41$    & $0.46$   &         &          &         &   \\
\\
\textbf{ActiveLab}     & $\bold{0.03}$    & $\bold{0.00}$    & $\bold{0.01}$    & $\bold{0.02}$   & $\bold{0.01}$  &          &         &   \\
\\
\textbf{Incremental}   & $\bold{0.01}$    & $\bold{0.01}$    & $\bold{0.01}$    & $\bold{0.01}$   & $\bold{0.00}$  & $0.31$   &         &   \\
\\
\textbf{ImpOr}            & $\bold{0.04}$    & $\bold{0.01}$    & $\bold{0.01}$    & $\bold{0.02}$   & $\bold{0.01}$  & $0.41$   & $0.37$  &   \\
\bottomrule
\end{tabular}
}
\caption{P-value matrix: AG News: Budget $50$.}
\label{tab_pvalue_ag_50}
\end{table}

\begin{comment}
\begin{table}[htbp]
\centering
\caption{P-value matrix: AG News: Budget $200$.}
\scriptsize
\renewcommand{\arraystretch}{1.3}
\resizebox{0.48\textwidth}{!}{
\begin{tabular}{|c|c|c|c|c|c|c|c|c|}
\hline
\textbf{} 
& \textbf{RR} 
& \textbf{ER} 
& \textbf{CR} 
& \textbf{BR} 
& \textbf{CEAL} 
& \textbf{ActiveLab} 
& \textbf{Incremental} 
& \textbf{ImpOr} \\
\hline
\textbf{RR} 
&  &  &  &  &  &  &  &  \\
\hline
\textbf{ER} 
& $\bold{0.00}$ &  &  &  &  &  &  &  \\
\hline
\textbf{CR} 
& $\bold{0.00}$ & $\bold{0.00}$ &  &  &  &  &  &  \\
\hline
\textbf{BR} 
& $0.19$ & $\bold{0.00}$ & $\bold{0.00}$ &  &  &  &  &  \\
\hline
\textbf{CEAL} 
& $\bold{0.01}$ & $\bold{0.00}$ & $0.27$ & $\bold{0.04}$ &  &  &  &  \\
\hline
\textbf{ActiveLab} 
& $\bold{0.00}$ & $\bold{0.00}$ & $\bold{0.00}$ & $\bold{0.00}$ & $\bold{0.00}$ &  &  &  \\
\hline
\textbf{Incremental} 
& $\bold{0.00}$ & $\bold{0.00}$ & $\bold{0.05}$ & $\bold{0.00}$ & $\bold{0.04}$ & $\bold{0.01}$ &  &  \\
\hline
\textbf{ImpOr} 
& $\bold{0.00}$ & $\bold{0.00}$ & $\bold{0.00}$ & $\bold{0.00}$ & $\bold{0.00}$ & $0.21$ & $\bold{0.00}$ &  \\
\hline
\end{tabular}
}
\label{tab_pvalue_ag_200}
\end{table}
\end{comment}

\begin{table}[ht]
\centering
\scriptsize
\resizebox{0.48\textwidth}{!}{
\begin{tabular}{lcccccccc}
\toprule
\textbf{} & \textbf{RR} & \textbf{ER} & \textbf{CR} & \textbf{BR} & \textbf{CEAL} & \textbf{ActiveLab} & \textbf{Incremental} & \textbf{ImpOr}\\
%\midrule
\textbf{RR}        &         &         &         &        &         &         &        \\
\\
\textbf{ER}       & $\bold{0.00}$    &           &          &         &          &         &   \\
\\
\textbf{CR}       & $\bold{0.00}$  & $\bold{0.00}$  &         &         &        &         &         &        \\
\\
\textbf{BR}         & $0.19$  & $\bold{0.00}$  & $\bold{0.00}$  &         &        &         &         &        \\
\\
\textbf{CEAL}          & $\bold{0.01}$  & $\bold{0.00}$  & $0.27$  & $\bold{0.04}$  &        &         &         &        \\
\\
\textbf{ActiveLab}     & $\bold{0.00}$  & $\bold{0.00}$  & $\bold{0.00}$  & $\bold{0.00}$  & $\bold{0.00}$ &         &         &        \\
\\
\textbf{Incremental}   & $\bold{0.00}$  & $\bold{0.00}$  & $\bold{0.05}$  & $\bold{0.00}$  & $\bold{0.04}$ & $\bold{0.01}$  &         &        \\
\\
\textbf{ImpOr}            & $\bold{0.00}$  & $\bold{0.00}$  & $\bold{0.00}$  & $\bold{0.00}$  & $\bold{0.00}$ & $0.21$  & $\bold{0.00}$  &        \\
\bottomrule
\end{tabular}
}
\caption{P-value matrix: AG News: Budget $200$}
\label{tab_pvalue_ag_200}
\end{table}

\begin{comment}
\begin{table}[htbp]
\centering
\caption{P-value matrix: AG News: Budget $300$.}
\scriptsize
\renewcommand{\arraystretch}{1.3}
\resizebox{0.48\textwidth}{!}{
\begin{tabular}{|c|c|c|c|c|c|c|c|c|}
\hline
\textbf{} 
& \textbf{RR} 
& \textbf{ER} 
& \textbf{CR} 
& \textbf{BR} 
& \textbf{CEAL} 
& \textbf{ActiveLab} 
& \textbf{Incremental} 
& \textbf{ImpOr} \\
\hline
\textbf{RR} 
&  &  &  &  &  &  &  &  \\
\hline
\textbf{ER} 
& $\bold{0.01}$ &  &  &  &  &  &  &  \\
\hline
\textbf{CR} 
& $\bold{0.02}$ & $\bold{0.01}$ &  &  &  &  &  &  \\
\hline
\textbf{BR} 
& $\bold{0.01}$ & $\bold{0.00}$ & $0.23$ &  &  &  &  &  \\
\hline
\textbf{CEAL} 
& $0.45$ & $\bold{0.00}$ & $0.07$ & $\bold{0.01}$ &  &  &  &  \\
\hline
\textbf{ActiveLab} 
& $\bold{0.00}$ & $\bold{0.00}$ & $\bold{0.00}$ & $\bold{0.00}$ & $\bold{0.00}$ &  &  &  \\
\hline
\textbf{Incremental} 
& $\bold{0.00}$ & $\bold{0.00}$ & $\bold{0.00}$ & $\bold{0.00}$ & $\bold{0.01}$ & $0.17$ &  &  \\
\hline
\textbf{ImpOr} 
& $\bold{0.00}$ & $\bold{0.00}$ & $\bold{0.00}$ & $\bold{0.00}$ & $\bold{0.01}$ & $0.29$ & $0.31$ &  \\
\hline
\end{tabular}
}
\label{tab_pvalue_ag_300}
\end{table}
\end{comment}

\begin{table}[ht]
\centering
\scriptsize
\resizebox{0.48\textwidth}{!}{
\begin{tabular}{lcccccccc}
\toprule
\textbf{} & \textbf{RR} & \textbf{ER} & \textbf{CR} & \textbf{BR} & \textbf{CEAL} & \textbf{ActiveLab} & \textbf{Incremental} & \textbf{ImpOr}\\
%\midrule
\textbf{RR}        &         &         &         &        &         &         &        \\
\\
\textbf{ER}       & $\bold{0.01}$    &           &          &         &          &         &   \\
\\
\textbf{CR}       & $\bold{0.02}$  & $\bold{0.01}$  &         &         &        &         &         &        \\
\\
\textbf{BR}         & $\bold{0.01}$  & $\bold{0.00}$  & $0.23$  &         &        &         &         &        \\
\\
\textbf{CEAL}          & $0.45$  & $\bold{0.00}$  & $0.07$  & $\bold{0.01}$  &        &         &         &        \\
\\
\textbf{ActiveLab}     & $\bold{0.00}$  & $\bold{0.00}$  & $\bold{0.00}$  & $\bold{0.00}$  & $\bold{0.00}$ &         &         &        \\
\\
\textbf{Incremental}   & $\bold{0.00}$  & $\bold{0.00}$  & $\bold{0.00}$  & $\bold{0.00}$  & $\bold{0.01}$ & $0.17$  &         &        \\
\\
\textbf{ImpOr}            & $\bold{0.00}$  & $\bold{0.00}$  & $\bold{0.00}$  & $\bold{0.00}$  & $\bold{0.01}$ & $0.29$  & $0.31$  &        \\
\bottomrule
\end{tabular}
}
\caption{P-value matrix: AG News: Budget $300$}
\label{tab_pvalue_ag_300}
\end{table}

\subsection{Study of Labeling Budget}

In this experiment, we studied the effect of the query budget per AL iteration. The results on the AG News dataset, for budgets $50, 200$ and $300$, are depicted in Figure \ref{fig_budget_results} (the budget for the results in Figure \ref{fig_agnews} was $100$); the corresponding p-value matrices are depicted in Tables \ref{tab_pvalue_ag_50}, \ref{tab_pvalue_ag_200} and \ref{tab_pvalue_ag_300} respectively. We notice a similar trend as in Figure \ref{fig_agnews}, where \textit{ActiveLab, Incremental} and \textit{ImpOr} outperform the other baselines. The improvement in performance achieved by these methods over \textit{RR, ER, CR, BR} and \textit{CEAL} is statistically significant across all the budgets $(p < 0.05)$. Among these three methods, there is no significant difference in performance for budgets $50$ and $300$, although \textit{ActiveLab} and \textit{ImpOr} outperform \textit{Incremental} at a significant level for budget $200$. These results reinforce our finding that AL methods designed to query informative samples in the presence of noisy annotators depict better performance than methods designed to query informative samples only. Further, AL methods which relabel a queried instance multiple times to ascertain the label (\textit{ActiveLab} and \textit{Incremental}) tend to outperform methods that attempt to ascertain the label based on the prediction from a single best annotator (such as \textit{CEAL}). The \textit{ER} method depicts poor performance (particularly, for budgets $200$ and $300$). 

\subsection{Performance using Machine Learning Models as Annotators}
\label{subsec_ML_Annotators}

In this experiment, we study the performance of AL algorithms using machine learning (ML) models as annotators (rather than human annotators), as conventionally done in active learning research. We trained $10$ machine learning models (\textit{Random Forest, Gradient Boosting, AdaBoost, SVM, Logistic Regression, XGBoost, Decision Tree, CatBoost, KMeans, ExtraTrees}) on the initial training set; their accuracies ranged from $56.27\%$ to $82.78\%$. When an unlabeled sample was queried, it was passed to the corresponding annotator (ML model) and the prediction of the model was used as the label of the unlabeled sample to retrain the underlying deep neural network. 

\begin{figure}[ht]
	\centering
         % \label{fig_agnews_GPT}
          \includegraphics[trim = 1.3in 3.2in 1.2in 3.4in,clip,width=.4\textwidth]{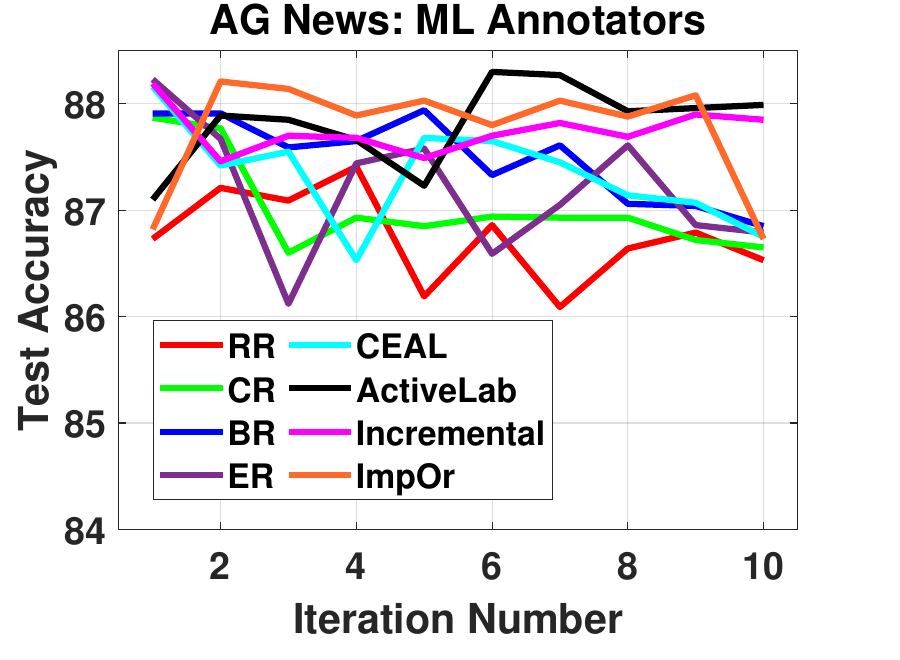}    
          \caption{Performance of AL algorithms using Machine Learning Models as Annotators on AG News dataset. Best viewed in color.}
        \label{fig_ML_Annotators_results}     
\end{figure} 

The results on the AG News dataset are depicted in Figure \ref{fig_ML_Annotators_results}. We note that using ML models as annotators, the accuracy of most of the methods remains more or less constant, or shows a decreasing trend. The accuracy values are also lower compared to the experiment when humans were used as annotators (Figure \ref{fig_agnews}). The accuracy of the \textit{ImpOr} method, for instance, starts at $\sim$$88.6\%$ and exceeds $89\%$ when humans are used as annotators (as evident from Figure \ref{fig_agnews}). With ML models as annotators, however, the accuracy starts at $\sim$$87\%$ and reaches a maximum value of $88.2\%$. The same applies for the \textit{ActiveLab} method, whose accuracy also exceeds $89\%$ with human annotators; but it attains the highest accuracy of $88.3\%$ when ML models are used as annotators. This shows that using ML models as annotators can degrade the performance of AL algorithms; the results also corroborate the necessity of developing intelligent AL algorithms designed to work with imperfect human annotators, rather than relying on ML models as annotators.

\section{Conclusion}
\label{sec_conc}

The goal of this paper was to study the performance of active learning algorithms under real-world annotation challenges. To this end, we collected crowd-sourced annotations on $3$ standard text classification datasets; we then conducted extensive experiments to study the performance of $8$ AL algorithms using the collected annotations. To the best of our knowledge, this is the first research effort to study the performance of deep active learning algorithms, on multi-class classification problems, with real-world crowd-sourced annotations. We hope this research will be a step toward bridging the gap between AL and real-world annotation challenges, and the insights gained will be useful for the deployment of AL systems in real-world applications. 

%\section{Limitations}
%\label{sec_limitations}

%Although our analysis has presented some interesting insights, it has some limitations. First, for some experiments, the performance of certain active learning algorithms could not be explained. For instance, the poor performance of \textit{CEAL} for the AG News dataset (Figure \ref{fig_agnews}) and \textit{ImpOr} for the Wikipedia dataset (Figure \ref{fig_wiki}), even though both these methods attempt to select the optimal annotator for each queried instance. These need further exploration and analysis. Moreover, we studied the performance of the algorithms on three datasets with relatively small number of classes: AG News ($4$ classes), Consumer Complaints ($6$ classes) and Wikipedia Movie Plots ($4$ classes). Future investigations should extend the analysis to datasets that contain more classes, multi-label datasets and hierarchical label datasets, which are common in text mining applications. 


\bibliographystyle{IEEEtran}
\bibliography{IJCNN2026}

\begin{thebibliography}{10}
\providecommand{\url}[1]{#1}
\csname url@samestyle\endcsname
\providecommand{\newblock}{\relax}
\providecommand{\bibinfo}[2]{#2}
\providecommand{\BIBentrySTDinterwordspacing}{\spaceskip=0pt\relax}
\providecommand{\BIBentryALTinterwordstretchfactor}{4}
\providecommand{\BIBentryALTinterwordspacing}{\spaceskip=\fontdimen2\font plus
\BIBentryALTinterwordstretchfactor\fontdimen3\font minus \fontdimen4\font\relax}
\providecommand{\BIBforeignlanguage}[2]{{%
\expandafter\ifx\csname l@#1\endcsname\relax
\typeout{** WARNING: IEEEtran.bst: No hyphenation pattern has been}%
\typeout{** loaded for the language `#1'. Using the pattern for}%
\typeout{** the default language instead.}%
\else
\language=\csname l@#1\endcsname
\fi
#2}}
\providecommand{\BIBdecl}{\relax}
\BIBdecl

\bibitem{Settles_2010}
B.~Settles, ``Active learning literature survey,'' in \emph{Technical Report 1648, University of Wisconsin-Madison}, 2010.

\bibitem{tong_support_2000}
S.~Tong and D.~Koller, ``Support vector machine active learning with applications to text classification,'' \emph{Journal of Machine Learning Research (JMLR)}, vol.~2, pp. 45--66, 2001.

\bibitem{Yoo_2019}
D.~Yoo and I.~Kweon, ``Learning loss for active learning,'' in \emph{IEEE Conference on Computer Vision and Pattern Recognition (CVPR)}, 2019.

\bibitem{Hatice_2010}
H.~Osmanbeyoglu, J.~Wehner, J.~Carbonell, and M.~Ganapathiraju, ``Active machine learning for transmembrane helix prediction,'' \emph{BMC Bioinformatics}, vol.~11, no.~1, 2010.

\bibitem{Pimentel_2020}
T.~Pimentel, M.~Monteiro, A.~Veloso, and N.~Ziviani, ``Deep active learning for anomaly detection,'' in \emph{IEEE International Joint Conference on Neural Networks (IJCNN)}, 2020.

\bibitem{Huang_2017}
S.~Huang, J.~Chen, X.~Mu, and Z.~Zhou, ``Cost-effective active learning from diverse labelers,'' in \emph{International Joint Conference on Artificial Intelligence (IJCAI)}, 2017.

\bibitem{Zhang_2015}
C.~Zhang and K.~Chaudhuri, ``Active learning from weak and strong labelers,'' in \emph{Neural Information Processing Systems (NIPS)}, 2015.

\bibitem{Yan_2016}
S.~Yan, K.~Chaudhuri, and T.~Javidi, ``Active learning from imperfect labelers,'' in \emph{Neural Information Processing Systems (NIPS)}, 2016.

\bibitem{Chakraborty_2020}
S.~Chakraborty, ``Asking the right questions to the right users: Active learning with imperfect oracles,'' in \emph{AAAI Conference on Artificial Intelligence}, 2020.

\bibitem{Yan_2011}
Y.~Yan, G.~Fung, R.~Rosales, and J.~Dy, ``Active learning from crowds,'' in \emph{International Conference on Machine Learning (ICML)}, 2011.

\bibitem{Ren_2021}
P.~Ren, Y.~Xiao, X.~Chang, P.~Huang, Z.~Li, B.~Gupta, X.~Chen, and X.~Wang, ``A survey of deep active learning,'' \emph{ACM Computing Surveys}, vol.~54, no.~9, 2021.

\bibitem{Khosla_2023}
S.~Khosla, C.~Whye, J.~Ash, C.~Zhang, K.~Kawaguchi, and A.~Lamb, ``Neural active learning on heteroskedastic distributions,'' in \emph{arXiv:2211.00928v2}, 2023.

\bibitem{Nuggehalli_2024}
S.~Nuggehalli, J.~Zhang, L.~Jain, and R.~Nowak, ``Direct: Deep active learning under imbalance and label noise,'' in \emph{arXiv:2312.09196v3}, 2024.

\bibitem{Shafir_2025}
N.~Shafir, G.~Hacohen, and D.~Weinshall, ``Active learning with a noisy annotator,'' in \emph{arXiv:2504.04506v1}, 2025.

\bibitem{Coreset_Paper}
O.~Sener and S.~Savarese, ``Active learning for convolutional neural networks: A core-set approach,'' in \emph{International Conference on Learning Representations (ICLR)}, 2018.

\bibitem{Badge_Paper}
J.~Ash, C.~Zhang, A.~Krishnamurthy, J.~Langford, and A.~Agarwal, ``Deep batch active learning by diverse, uncertain gradient lower bounds,'' in \emph{International Conference on Learning Representations (ICLR)}, 2020.

\bibitem{TOD_ICCV_2021}
S.~Huang, T.~Wang, H.~Xiong, J.~Huan, and D.~Dou, ``Semi-supervised active learning with temporal output discrepancy,'' in \emph{IEEE International Conference on Computer Vision (ICCV)}, 2021.

\bibitem{liu2021influence}
Z.~Liu, H.~Ding, H.~Zhong, W.~Li, J.~Dai, and C.~He, ``Influence selection for active learning,'' in \emph{IEEE International Conference on Computer Vision (ICCV)}, 2021.

\bibitem{Sinha_2019}
S.~Sinha, S.~Ebrahimi, and T.~Darrell, ``Variational adversarial active learning,'' in \emph{IEEE International Conference on Computer Vision (ICCV)}, 2019.

\bibitem{Zhu_2017}
J.~Zhu and J.~Bento, ``Generative adversarial active learning,'' in \emph{arXiv:1702.07956}, 2017.

\bibitem{Ducoffe_2018}
M.~Ducoffe and F.~Precioso, ``Adversarial active learning for deep networks: a margin based approach,'' in \emph{International Conference on Machine Learning (ICML)}, 2018.

\bibitem{Chen_2022}
Y.~Chen, K.~Sankararaman, A.~Lazaric, M.~Pirotta, D.~Karamshuk, Q.~Wang, K.~Mandyam, S.~Wang, and H.~Fang, ``Improved adaptive algorithm for scalable active learning with weak labeler,'' in \emph{arXiv:2211.02233v1}, 2022.

\bibitem{Donmez_2008}
P.~Donmez and J.~Carbonell, ``Proactive learning: cost-sensitive active learning with multiple imperfect oracles,'' in \emph{ACM Conference on Information and Knowledge Management (CIKM)}, 2008.

\bibitem{Donmez_2009}
P.~Donmez, J.~Carbonell, and J.~Schneider, ``Efficiently learning the accuracy of labeling sources for selective sampling,'' in \emph{ACM Conference on Knowledge Discovery and Data Mining (KDD)}, 2009.

\bibitem{Zheng_2010}
Y.~Zheng, S.~Scott, and K.~Deng, ``Active learning from multiple noisy labelers with varied costs,'' in \emph{IEEE International Conference on Data Mining (ICDM)}, 2010.

\bibitem{Ipeirotis_2014}
P.~Ipeirotis, F.~Provost, V.~Sheng, and J.~Wang, ``Repeated labeling using multiple noisy labelers,'' \emph{Data Mining and Knowledge Discovery}, vol.~28, 2014.

\bibitem{Zhao_2011}
L.~Zhao, G.~Sukthankar, and R.~Sukthankar, ``Incremental relabeling for active learning with noisy crowdsourced annotations,'' in \emph{International Conference on Social Computing}, 2011.

\bibitem{Goh_2023}
H.~Goh and J.~Mueller, ``Activelab: Active learning with re-labeling by multiple annotators,'' in \emph{International Conference on Learning Representations Workshop (ICLR-W)}, 2023.

\bibitem{Yan_2012}
Y.~Yan, R.~Rosales, G.~Fung, F.~Farooq, B.~Rao, and J.~Dy, ``Active learning from multiple knowledge sources,'' in \emph{International Conference on Artificial Intelligence and Statistics (AISTATS)}, 2012.

\bibitem{AGNews_dataset}
X.~Zhang, J.~Zhao, and Y.~LeCun, ``Character-level convolutional networks for text classification,'' in \emph{Neural Information Processing Systems (NeurIPS)}, 2015.

\bibitem{wang_probability_2008}
Q.~A. Wang, ``Probability distribution and entropy as a measure of uncertainty,'' \emph{Journal of Physics A: Mathematical and Theoretical}, vol.~41, 2008.

\bibitem{BERT_paper}
J.~Devlin, M.~Chang, K.~Lee, and K.~Toutanova, ``{BERT}: Pre-training of deep bidirectional transformers for language understanding,'' in \emph{Nations of the Americas Chapter of the Association for Computational Linguistics (NAACL)}, 2019.

\bibitem{AL_Stopping_1}
Y.~Zhang, W.~Cai, W.~Wang, and Y.~Zhang, ``Stopping criterion for active learning with model stability,'' \emph{ACM Transactions on Intelligent Systems and Technology (TIST)}, vol.~9, 2017.

\bibitem{AL_Stopping_2}
J.~Zhu, H.~Wang, and E.~Hovy, ``Learning a stopping criterion for active learning for word sense disambiguation and text classification,'' in \emph{International Joint Conference on Natural Language Processing (IJNLP)}, 2008.

\bibitem{AL_Stopping_3}
H.~Ishibashi and H.~Hino, ``Stopping criterion for active learning based on deterministic generalization bounds,'' in \emph{International Conference on Artificial Intelligence and Statistics (AISTATS)}, 2020.

\bibitem{Xue_2019}
Y.~Xue and M.~Hauskrecht, ``Active learning of multi-class classification models from ordered class sets,'' in \emph{AAAI Conference on Artificial Intelligence}, 2019.

\bibitem{Biology_paper}
A.~Rzhetsky, H.~Shatkay, and W.~Wilbur, ``How to get the most out of your curation effort,'' \emph{PLoS Computational Biology}, vol.~5, no.~5, 2009.

\bibitem{GPT2_Paper}
A.~Radford, J.~Wu, R.~Child, D.~Luan, D.~Amodei, and I.~Sutskever, ``Language models are unsupervised multitask learners,'' in \emph{Technical Report, OpenAI}, 2019.

\end{thebibliography}

%\begin{comment}

\newpage

\appendix

\section{Appendix}
\label{sec_appendix}

We present the following in this Appendix. 

\begin{itemize}

\item Performance on Scientific Text Data (Section \ref{subsec_Biology})

\item Performance using a Subset of Annotators (Section \ref{subsec_3annotators})

\item Error bar plots (Section \ref{subsec_errorbar})

\item Further analysis of the crowd-sourced annotations (Section \ref{subsec_analysis})

\item Sample crowd-sourced annotations for the three datasets (Section \ref{subsec_sample_ann})

\item Full Data Collection Protocol (Section \ref{subsec_protocol}) 

\item Performance Analysis with the GPT-2 Backbone (Section \ref{subsec_GPT2})

\end{itemize}

%\subsection {Implementation Details}
%\label{subsec_implementation}

%The BERT model was trained using the conventional cross entropy loss; we used a learning rate of $1e^{-5}$ (with cosine annealing), momentum of $0.9$, weight decay of $5e^{-7}$, and the \textit{Adam} optimizer. The experiments were performed on a workstation with 64 GB RAM and two NVIDIA Quadro RTX 5000 GPUs with 16 GB memory each. The implementations were performed using PyTorch; the Python packages used include Hugging Face Transformers for BERT/Bert Tokenizer, Scikit-Learn, NumPy, Pandas, and Scipy.

\begin{figure*}[htbp]
    \centering
    \subfloat[Polarity Label]{%
        \label{fig_biology_polarity}
        \includegraphics[
            trim=1.3in 3.2in 1.7in 3.4in,
            clip,
            width=0.35\linewidth
        ]{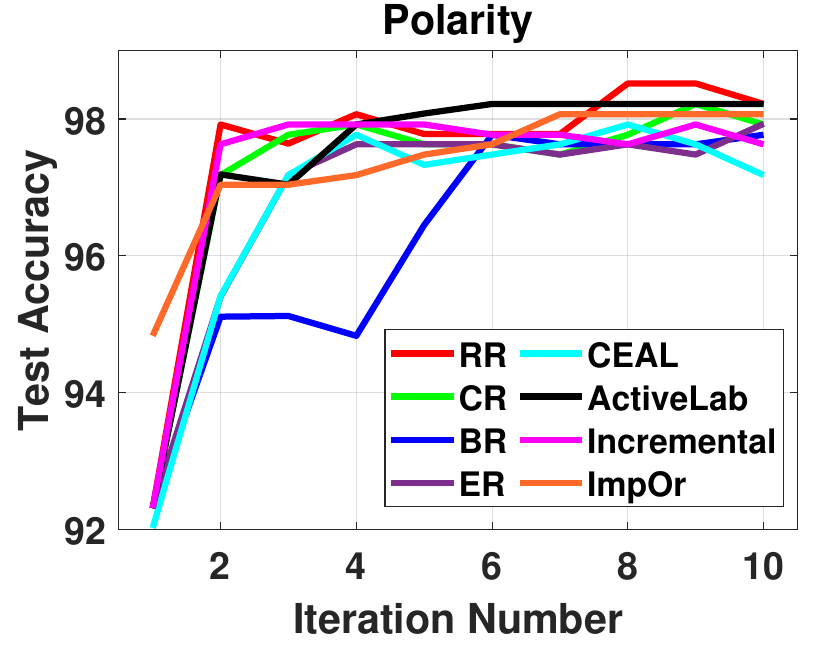}
    }
    \hspace{.2in}
    \subfloat[Evidence Label]{%
        \label{fig_biology_evidence}
        \includegraphics[
            trim=1.3in 3.2in 1.7in 3.4in,
            clip,
            width=0.35\linewidth
        ]{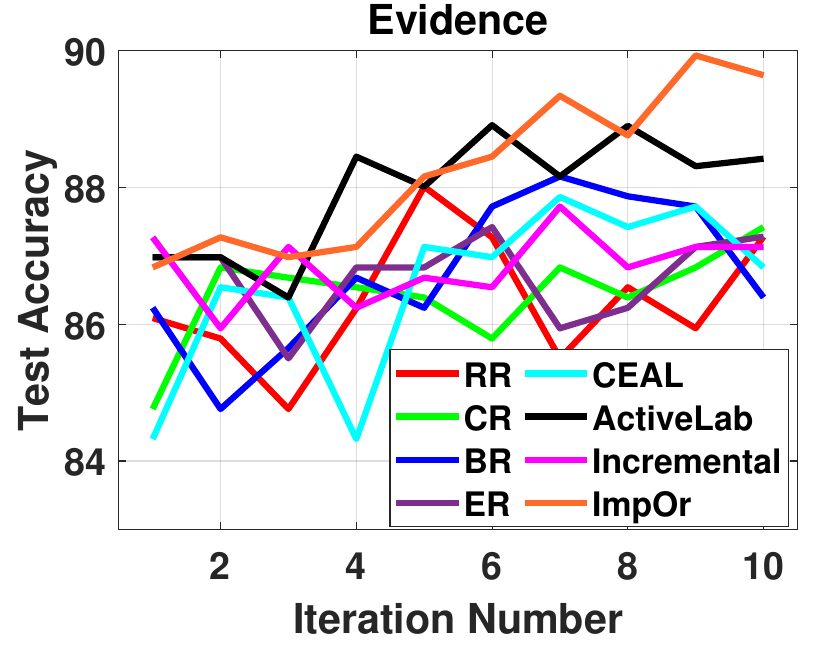}
    }
     \caption{Performance of AL algorithms on scientific text data (PubMed and GeneWays corpus).}
    \label{fig_biology_results}
\end{figure*}

\subsection{Performance on Scientific Text Data}
\label{subsec_Biology}

The goal of this experiment was to study the performance of the active learning algorithms on a scientific text dataset. We used the PubMed and GeneWays corpus made publicly available by \cite{Biology_paper}. It contains a corpus of $10,000$ sentences each that has been annotated by $3$ out of $8$ available annotators. Each sentence has been annotated with several labels. We used the \textit{polarity} and \textit{evidence} labels in our study. As proposed in \cite{Yan_2012}, we binarized them into two classes, containing positive and negative labels. We used a subset of $1,000$ examples, where each sentence was labeled by five annotators. We used $300$ samples as the initial training set, $400$ samples as the unlabeled set and the remaining samples for testing. 

The results are depicted in Figure \ref{fig_biology_results}; they are consistent with the conclusions drawn in Section \ref{subsec_AL_perf}, that AL algorithms which are designed to operate with imperfect annotators generally outperform methods that assume oracles to be infallible and attempt to query the informative unlabeled samples only. This is particularly true for the \textit{evidence} label, where \textit{ActiveLab} and \textit{ImpOr} depict the best AL performance.

\begin{figure*}[htbp]
    \centering
    \subfloat[Best Annotators]{%
        \label{fig_agnews_best_annotators}
        \includegraphics[
            trim=1.3in 3.2in 1.7in 3.4in,
            clip,
            width=0.31\linewidth
        ]{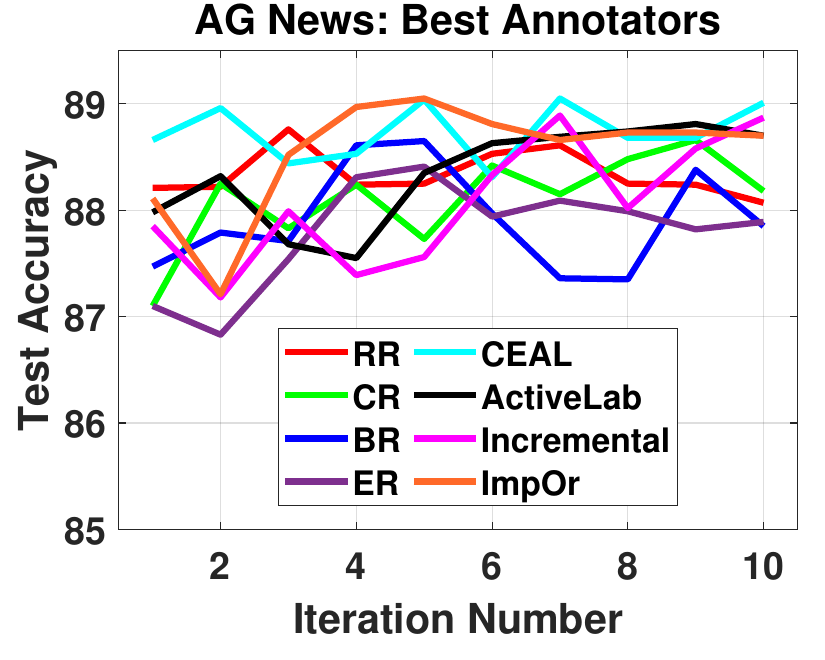}
    }
    \hfill
    \subfloat[Random Annotators]{%
        \label{fig_agnews_random_annotators}
        \includegraphics[
            trim=1.3in 3.2in 1.7in 3.4in,
            clip,
            width=0.31\linewidth
        ]{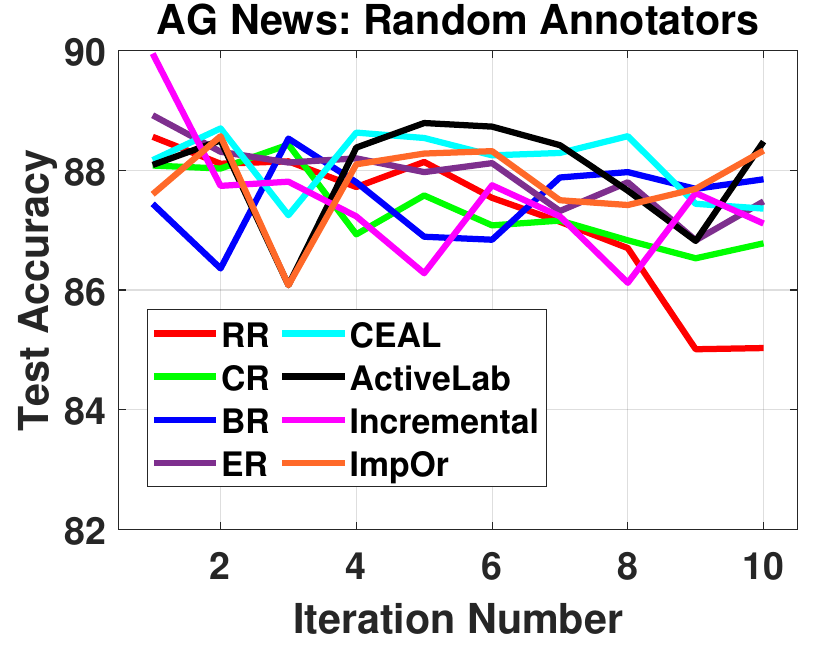}
    }
    \hfill
    \subfloat[Worst Annotators]{%
        \label{fig_agnews_worst_annotators}
        \includegraphics[
            trim=1.3in 3.2in 1.7in 3.4in,
            clip,
            width=0.31\linewidth
        ]{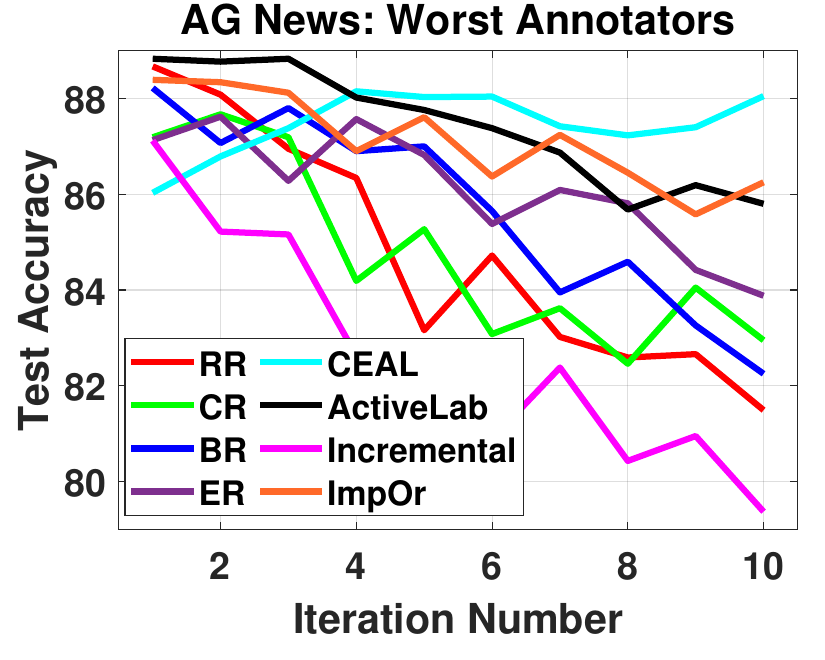}
    }
    \caption{Study of AL algorithms on the AG News dataset using a subset of annotators. Best viewed in color.}
    \label{fig_3annotators_results}
\end{figure*}

\subsection{Performance using a Subset of Annotators}
\label{subsec_3annotators}

In this experiment, we studied the performance of the AL algorithms using a subset of annotators, instead of all the $10$ annotators. We specifically studied three setups: $(i)$ using the $3$ best annotators (with the highest annotation accuracies); $(ii)$ using the $3$ worst annotators (with the lowest annotation accuracies); and $(iii)$ using $3$ random annotators. The results on the AG News dataset are depicted in Figure \ref{fig_3annotators_results}. The performance shows a decreasing trend as we move from the best to the worst annotators. This is expected, as with a subset of the worst annotators, most of the provided annotations will be incorrect, which will pollute the training set and degrade the model performance. 

Comparing Figure \ref{fig_agnews_best_annotators} (result with best annotators) with Figure \ref{fig_agnews} (result with all annotators), we make the following observations: $(i)$ For some algorithms (such as \textit{CEAL}), the AL performance improves significantly when only the best set of annotators are used. This is probably because \textit{CEAL} attempts to find the single best annotator for a given unlabeled sample; using the best set of annotators ensures that the obtained annotations will mostly be correct. Thus, for some AL algorithms, using a subset of only the best annotators may improve the learning performance, than using all the available annotators. \textit{CEAL} also shows the most robustness when the worst set of annotators are used. $(ii)$ For some algorithms (such as \textit{ActiveLab}), the AL performance degrades slightly when only the best set of annotators are used. \textit{ActiveLab} collates predictions from multiple annotators to derive the annotation of a queried unlabeled sample. If only the best set of annotators are used (with high annotation costs), a high price needs to be paid to get a single sample annotated, which limits the total number of unlabeled samples that can be queried under the given budget. When all the annotators are used, \textit{ActiveLab} can intelligently decide which annotators to select to annotate a given sample so as to optimize the annotation quality and the budget; this can potentially result in more sample queries, improving the learning performance. $(iii)$ All the methods show a rapid degradation in performance when the worst set of annotators are used, which is intuitive.

\subsection{Error bar Plots}
\label{subsec_errorbar}

\begin{figure*}[htbp]
    \centering
    \subfloat[AG News]{%
        \label{fig_agnews_errorbar}
        \includegraphics[
            trim=1.3in 3.2in 1.7in 3.4in,
            clip,
            width=0.31\linewidth
        ]{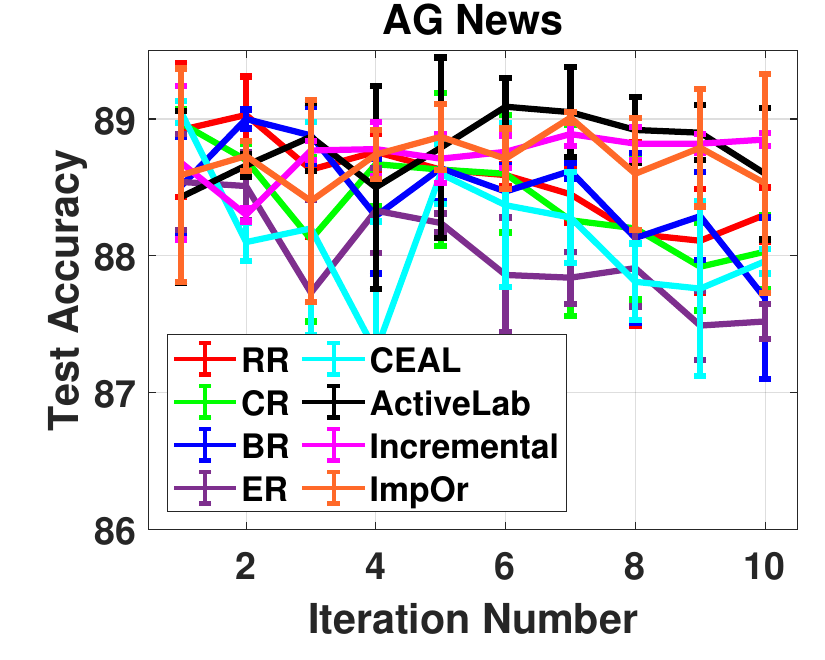}
    }
    \hfill
    \subfloat[Consumer Complaint]{%
        \label{fig_cc_errorbar}
        \includegraphics[
            trim=1.3in 3.2in 1.7in 3.4in,
            clip,
            width=0.31\linewidth
        ]{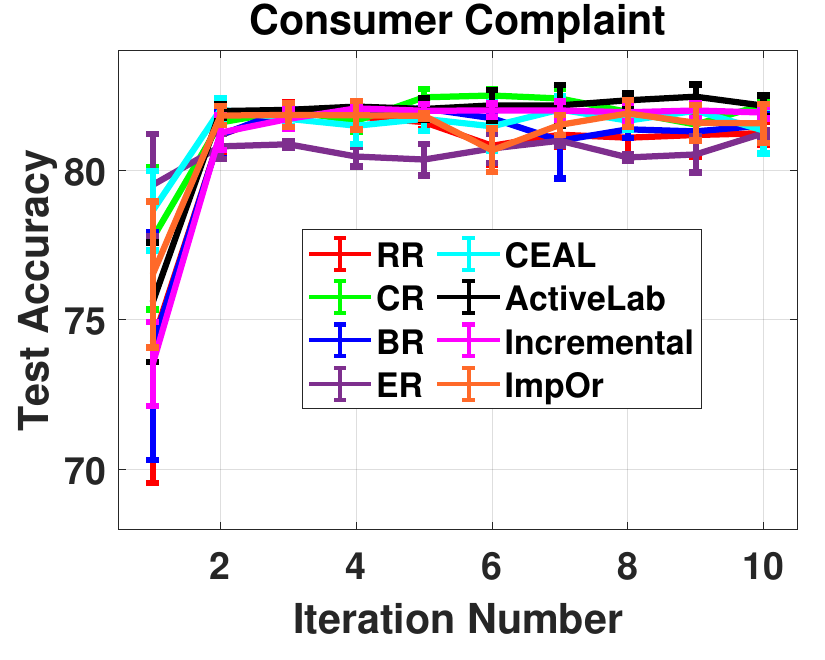}
    }
    \hfill
    \subfloat[Wikipedia Movie Plots]{%
        \label{fig_wiki_errorbar}
        \includegraphics[
            trim=1.3in 3.2in 1.7in 3.4in,
            clip,
            width=0.31\linewidth
        ]{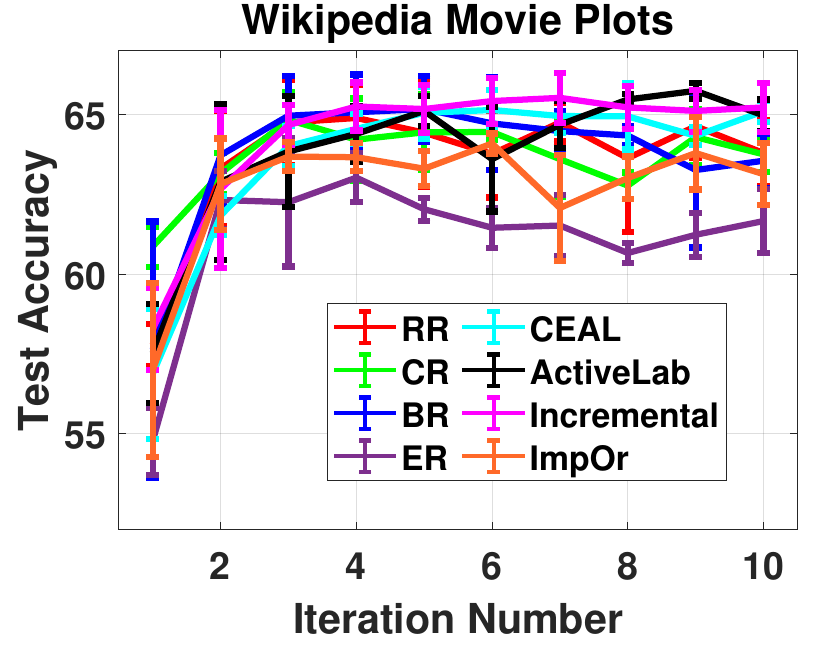}
    }
    \caption{Study of AL performance (with error bars). Best viewed in color.}
    \label{fig_main_results_errorbar}
\end{figure*}

\begin{figure*}[htbp]
    \centering
    \subfloat[Budget 50]{%
        \label{fig_agnews_budget_50_errorbar}
        \includegraphics[
            trim=1.3in 3.2in 1.7in 3.4in,
            clip,
            width=0.31\linewidth
        ]{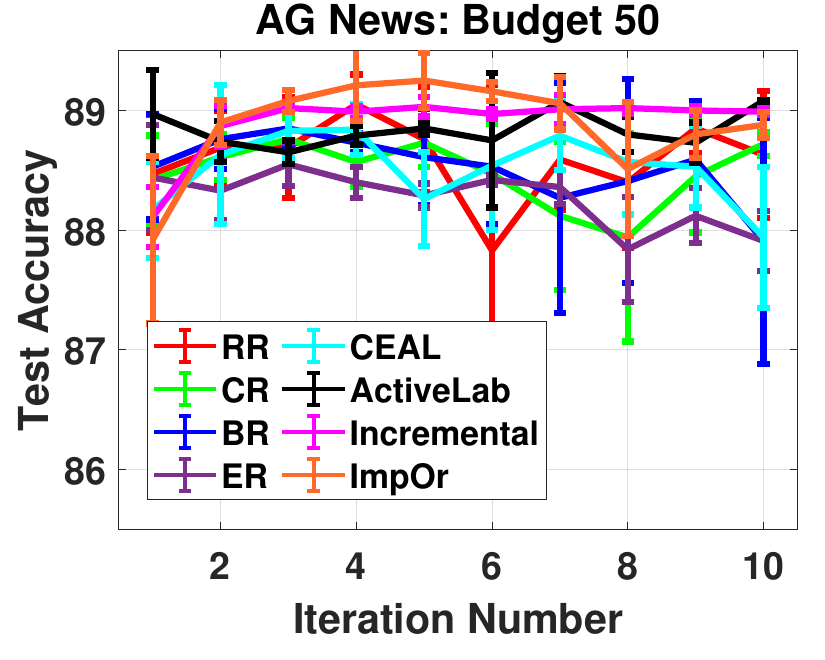}
    }
    \hfill
    \subfloat[Budget 200]{%
        \label{fig_agnews_budget_200_errorbar}
        \includegraphics[
            trim=1.3in 3.2in 1.7in 3.4in,
            clip,
            width=0.31\linewidth
        ]{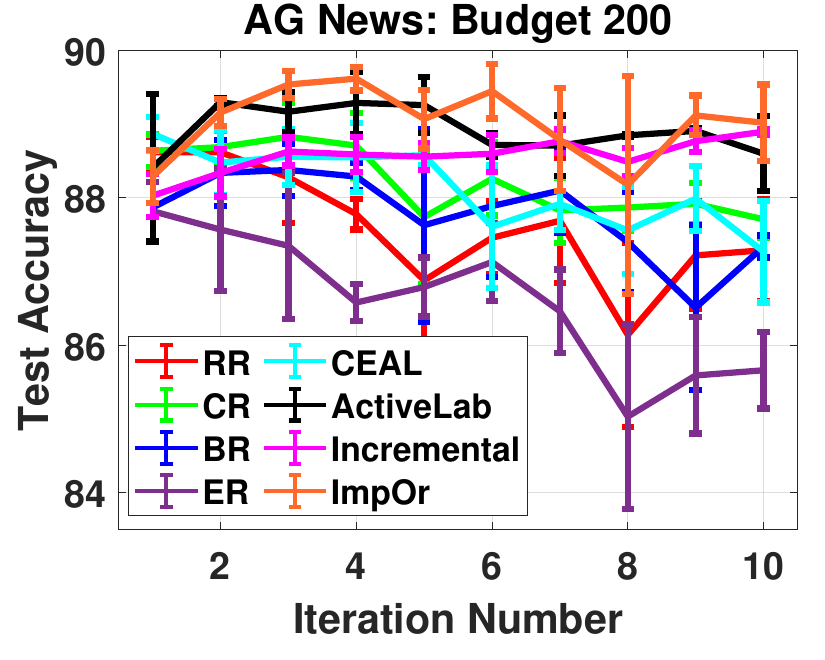}
    }
    \hfill
    \subfloat[Budget 300]{%
        \label{fig_agnews_budget_300_errorbar}
        \includegraphics[
            trim=1.3in 3.2in 1.7in 3.4in,
            clip,
            width=0.31\linewidth
        ]{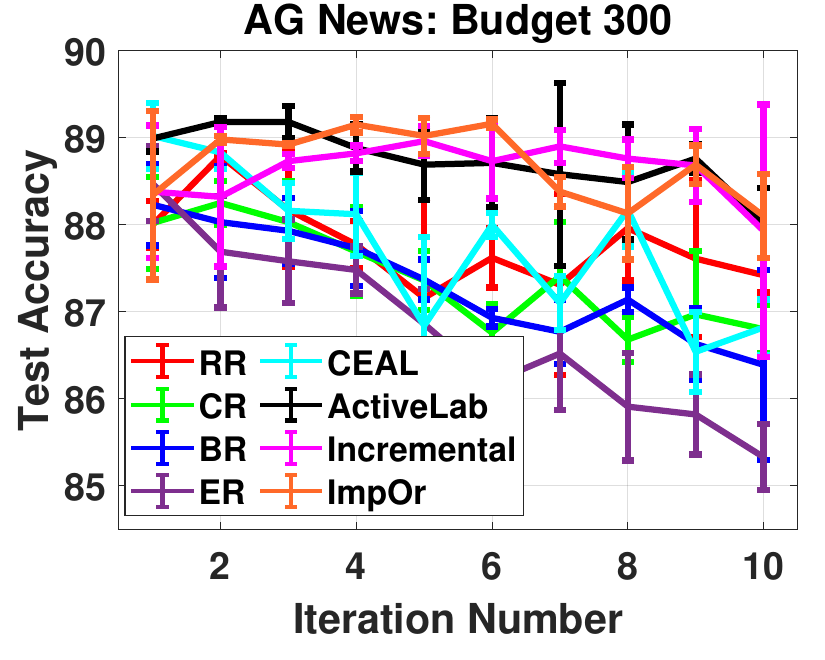}
    }
    \caption{Study of labeling budget on the AG News dataset (with error bars). Best viewed in color.}
    \label{fig_budget_results_errorbar}
\end{figure*}

In this section, we present the active learning graphs (from the main paper) with error bars. The AL performance results on the three datasets are depicted in Figure \ref{fig_main_results_errorbar}. The results with different query budgets are shown in Figure \ref{fig_budget_results_errorbar}. \textit{RR} and \textit{BR} depict a comparatively higher variance in the results (particularly for budgets $50$ and $200$). The other methods depict more or less stable performance across different datasets and query budgets; \textit{CEAL} and \textit{ImpOr} depict a relatively higher variance for the AG News experiment (Figure \ref{fig_agnews_errorbar}). 

\subsection{Further Analysis of the Crowd-sourced Annotations}
\label{subsec_analysis}

\begin{figure*}[ht]
    \centering
    \subfloat[AG News]{%
        \label{fig_agnews_confusion}
        \includegraphics[
            width=0.31\linewidth
        ]{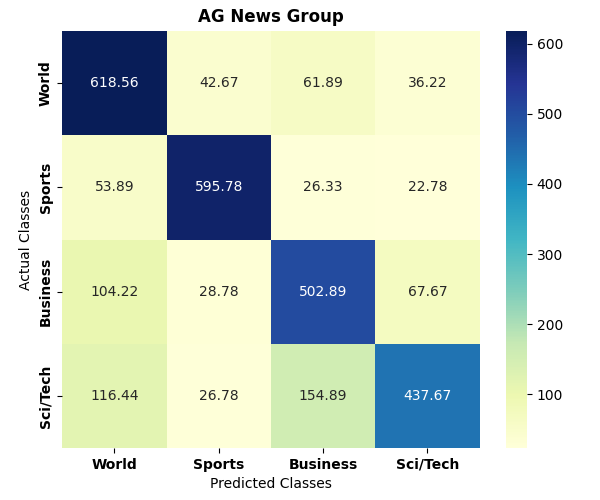}
    }
    \hfill
    \subfloat[Consumer Complaint]{%
        \label{fig_cc_confusion}
        \includegraphics[
            width=0.31\linewidth
        ]{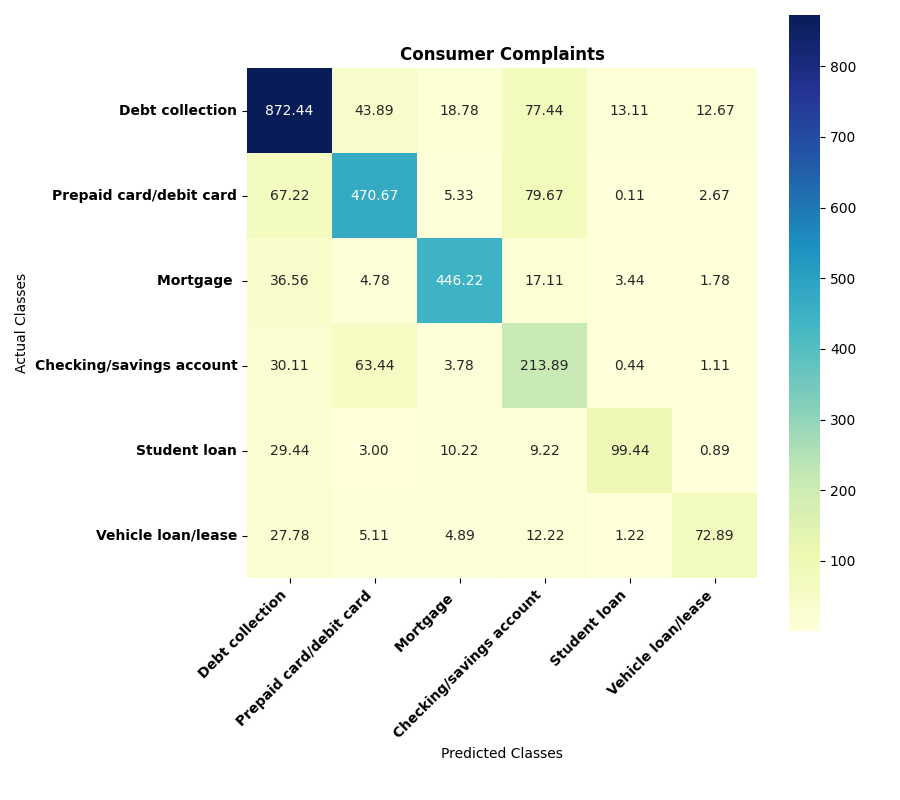}
    }
    \hfill
    \subfloat[Wikipedia Movie Plots]{%
        \label{fig_wiki_confusion}
        \includegraphics[
            width=0.31\linewidth
        ]{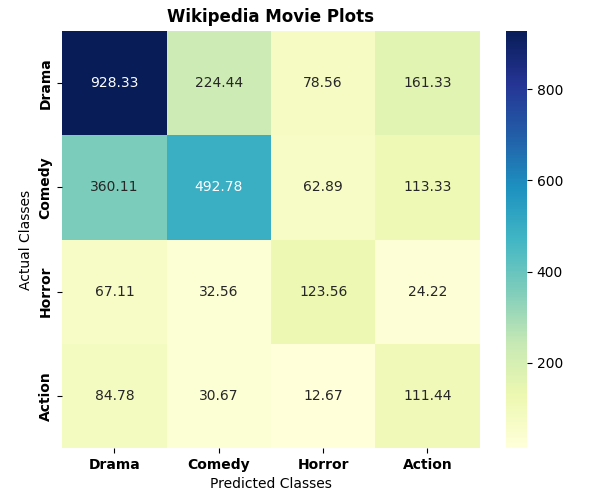}
    }
    \caption{Confusion matrices of the crowd-sourced annotations for each dataset (averaged across all the annotators). Best viewed in color.}
    \label{fig_confusion_matrices}
\end{figure*}

Figure \ref{fig_confusion_matrices} depicts the confusion matrices of the crowd-sourced annotations for each dataset, averaged across all the annotators. For the AG News dataset, the highest error was furnished by samples from class \textit{``Sci/Tech''} being annotated as belonging to class \textit{``Business''}. An example of such a sample is shown below: \\

\textit{``Reuters - Two British entrepreneurs who founded online dating service Udate.com and sold it last year to media mogul Barry Diller's InterActive Corp  for  36;150 million, have resurfaced to run a professionals-networking Web start-up.''} \\

For the Consumer Complaints dataset, the highest error was furnished by samples from class \textit{``Prepaid card / debit card''} being annotated as belonging to class \textit{``Checking / savings account''}. An example of such a sample is shown below: \\

\textit{``Since XX/XX/2020 Capital One has decided to start holding funds of which I now have to call them and they call my Bank to verify funds and then they make a exception to release the funds even though they took them out 3-5days prior. Now I have been on the phone for over 2.5hrs as they once again holding funds typical garbage 2-3x week call this time they verified the funds but did not put all the funds in not even half stating they are going to hold the funds for a extended amount 30-90 days to make sure no Fraud. I am the Owner of the Card and use this for my XXXX XXXX and been charging 1-2k a month for over 6 months and now since this year specifically its been nothing but a nightmare holds, customer service is terrible as they dont understand and dont care.''} \\

For the Wikipedia Movie Plots dataset, the highest error was furnished by samples from class \textit{``Comedy''} being annotated as belonging to class \textit{``Drama''}. An example of such a sample is shown below: \\

\textit{``Episodic in nature (effectively three short films merged into one), the first episode features Hubby winning a live turkey in a raffle and taking it home on a crowded streetcar, much to the chagrin of the other passengers. The second features Hubby grudgingly taking the family en masse out on his brand new Butterfly Six automobile, and the third is an escapade with his sleepwalking mother-in-law.''} \\

Some of these errors make intuitive sense, for instance, a sample from class \textit{``Prepaid card / debit card''} being annotated as belonging to class \textit{``Checking / savings account''}. As mentioned in Section 4.1 in the main paper, a possible solution may be to allow annotators to specify a \textit{second-best} preference, along with their top preference, when annotating a sample, and developing an AL algorithm accordingly. 

\subsection{Sample Annotations}
\label{subsec_sample_ann}

Sample crowd-sourced annotations from the AG News, Consumer Complaint and Wikipedia Movie Plots datasets are shown in Tables \ref{tab_ag_annotations}, \ref{tab_cc_annotations} and \ref{tab_wiki_annotations} respectively. 

\begin{table*}[htbp]
\centering
\caption{Sample annotations and ground truth values from the AG News Groups dataset. 
\textbf{Unknown} indicates that the annotator did not provide a label.}
\scriptsize
\renewcommand{\arraystretch}{1.15}
\resizebox{\textwidth}{!}{
\begin{tabular}{|p{6.2cm}|p{2cm}|p{5.6cm}|}
\hline
\textbf{Text} & \textbf{Ground Truth} & \textbf{Annotations} \\
\hline
No one had to remind Dallas and Minnesota about how little Sunday's opener at the Metrodome means over the course of a full season.
& Sports
& Unknown, Sports, Sports, Sports, Sports, Sports, Sci/Tech, Sports, Sports, Sports \\
\hline
Hakim Warrick scored 19 points, grabbed 10 rebounds and handed out five assists to lead No.~6 Syracuse to a 104--54 rout Thursday of Northern Colorado.
& Sports
& Sports, Sports, Sports, Sports, Sports, Sports, World, Sports, Sports, Sports \\
\hline
Currency traders, investors and strategists are more bearish on the dollar than at any time in the past 18 months, a Bloomberg News survey indicates.
& World
& Business, Business, Business, Business, Business, Business, Sports, Business, Business, Business \\
\hline
XSTRATA, the diversified miner, will have to increase its A\$7.4 billion bid for WMC Resources if it is to succeed, analysts have said.
& Business
& Business, Business, Business, Business, Business, Business, Business, Business, Business, Business \\
\hline
Sony's PlayStation console has been so popular that it feels strange that PSP is their first attempt at the handheld gaming market.
& Sci/Tech
& Sci/Tech, Sci/Tech, Sports, Sci/Tech, Sci/Tech, Sci/Tech, Business, Business, Business, Sci/Tech \\
\hline
\end{tabular}
}
\label{tab_ag_annotations}
\end{table*}

\begin{table*}[htbp]
\caption{Sample annotations and ground truth values from the Consumer Complaints dataset. Note: Here \textbf{Unknown} means the annotator did not provide a label for that sample.}
\centering
\small
\resizebox{\textwidth}{!}{%
\renewcommand{\arraystretch}{1.3}  % Optional: adds spacing between rows
\begin{tabular}{p{15cm}p{2cm}p{5cm}}
  \toprule
  \textbf{Text} & \textbf{Ground Truth} & \textbf{Annotations} \\
  \midrule

  \parbox{15cm}{I am attaching previously agreed and signed documentation from standing in front of the presiding judge on XX/XX/XXXX, XXXX For settlement due of {\$1300.00}, of 45 monthly payments of XXXX Then the following documents from XX/XX/XXXX, is whenever our signed agreement was ripped right out from underneath me by a Motion and Order to Set Aside Judgment without any warning. It states that I was served via First Class Mail to address XXXX XXXX XXXX, XXXX TX XXXX - when in fact I NEVER received those documents/were served, nor was I living there - I had moved from that address to another a year + prior to us had even settling on XX/XX/XXXX, XXXX Such happenings, I was never properly served, and there is no proof/signature ( by me ) proving that I was served the documents before the Motion and Order to Set Aside Judgment was submitted on XX/XX/XXXX then was granted on XX/XX/XXXX. There was no trial called, the only trial I was informed of was for XX/XX/XXXX XXXX which I had attended, due to us agreeing and signing in front of the judge ). Now, TWO years later, it is now sent off to another debt collector, XXXX XXXX , XXXX is calling attempting to collect the debt, however, they are giving me time in order to submit my complaints and attempt to dispute the amounts CACH LLC is trying to recover from me.} 
  & Debt collection 
  & \parbox{5cm}{Debt collection, Debt collection, Mortgage, Debt collection, Unknown, Debt collection, Debt collection, Debt collection, Debt collection, Debt collection}  \\ \\
  
  \parbox{15cm}{On XX/XX/2020 I received 2 phone calls from XXXX at my work number ( XXXX ) XXXX and 2 calls from ( XXXX from Automotive Credit Corp ) on my cell ( XXXX ). These all occur first thing in the morning when I am working. I returned the call around XXXX on my lunch and left a VM to let them know that they are not to call my work phone number ( XXXX ) as my boss got the message and was not happy. I was given a verbal warning. On XX/XX/2020 I again received 2 calls from XXXX ( XXXX from Automotive Credit Corp ) at my work number and also 2 calls from them on my cell. I then sent them an email which I got from their website and which I asked them again not to contact me at the work number. About an hour later a friend of mine who I listed on the loan papers as a reference received a call from XXXX ( XXXX ) who told XXXX I was behind in payment and that she needed a number to reach me at. XXXX then called me and asked that I remove her information from the file. I also requested that. Now on XX/XX/2020 I again receive 2 calls on my cell phone and one at my work number which my boss was lucky enough to get. He then gave me a final written warning and explained no more personal business calls at work. I have not included the emails sent to them or the write up as it may need to be used in court and I do not wish to provide them anything until after I speak to an Attorney.} 
  & Vehicle loan/lease 
  & Debt collection, Vehicle loan/lease, Unknown, Vehicle loan/lease, Unknown, Debt collection, Debt collection, Vehicle loan/lease, Student loan, Unknown \\ \\
  
  \parbox{15cm}{I set up a transfer to pull from by XXXX checking account to my People 's United Bank checking account. This was a hassle in and of itself, however, it worked - the transfer went through. A month later, another transaction went through for the same amount. I thought it must be a mistake, I don't recall setting this up to be recurring. So I logged into my online account and checked my "" Transfers '' tab. It read "" No Transfer Accounts Added. '' I checked my Bill Pay ; no recurring bill pay 's were set up. I checked every possible avenue and no scheduled payments or transfers were shown anywhere on my account. Despite this, today another transaction posted as pending on my XXXX account. Since I was not expecting this, it over-drafted my account. My communication with XXXX indicated it was initiated by People 's. So I called People 's and they would not stop the transaction when I requested to do so, even though it was still pending. People 's indicated that this recurring transfer was set up "" internally '' and that is why it didn't show up anywhere on my account. What does that even mean? Anyways, since I was forced to cancel the transaction through XXXX ( People 's would not oblige ) I was charged a {\$36.00} stoppage fee. This was all a nightmare and took more than an hour of my workday to resolve. It does not seem right that a recurring transfer can be ongoing with ZERO indication it exists anywhere in my online account.} 
  & Checking/savings account 
  & Prepaid card/debit card, Checking/savings account, Checking/savings account, Checking/savings account, Checking/savings account, Checking/savings account, Checking/savings account, Checking/savings account, Checking/savings account, Checking/savings account \\ \\
  
  \parbox{15cm}{We started with XXXX XXXX, then XXXX XXXX and now Selene Finance. We have submitted so many documents to XXXX XXXX, We submitted so many packages to Loan Care and thought we were in the final stages of getting a loan modification with a sizable down payment and then we were notified that the loan has been transferred to Selene Finance. Selene Finance does not want to hear anything about what was sent to XXXX XXXX  before they took over my account. I have also received a Summons \& Complaint to foreclose on my home. We submitted documents to XXXX XXXX on the following dates : XX/XX/XXXX XX/XX/XXXX XX/XX/XXXX XX/XX/XXXX XX/XX/XXXX XX/XX/XXXX XX/XX/XXXX We received a confirmation from XXXX XXXX Customer support acknowledgment from XXXX XXXX with Ticket  : XXXX Then more requests on XX/XX/XXXX, XX/XX/XXXX XX/XX/XXXX Missing documents and request for a new RMA. We have started over so many times with XXXX and now Selene claim they do not see my application from XXXX XXXX and I received a Summons and Complaint Please help!}
  & Mortgage 
  & Mortgage, Mortgage, Debt collection, Mortgage, Mortgage, Student loan, Mortgage, Mortgage, Mortgage, Mortgage \\ \\
  
  \parbox{15cm}{I have been a loyal customer with a prompt payment history with Barclay's/Juniper Mastercard. Back in XX/XX/2019, I noticed a pending charge on my Barclay 's Mastercard. I called the bank immediately and advised them this was not my charge. This charge was dated was XX/XX/XXXX, amount was for XXXX from XXXX ( I never even heard of this company ). They removed the charge from my account. I received a confirmation letter on dated XX/XX/XXXX advising an investigation would be made pending the outcome, I would not be held responsible.XXXX XXXX I received a second letter determining I am not responsible, pending the merchants response. On around XX/XX/XXXX, I received a call from an XXXX number, at first I didn't want to answer, it was Barclays saying they considered the transaction valid, claiming the package was delivered to my address. I closed the account immediately. Several days later, I received a letter saying the transaction was valid and the charge would be placed back on my account. As a result of these findings, I did reach XXXX and XXXX. The XXXX tracking information shows a package sent, and per the customer it was pulled off the truck and signed fraudulently at a XXXX facility. I also received a copy of the fraudulent signature. I reached out to XXXX, it seems to purchase was made online with a fake email address and a SECOND attempt was made but failed validity the following day on XX/XX/XXXX. I followed up with XXXX with a case number advising them of the fraud transaction. I gathered all the information along a police report and all attached documents from XXXX scanned and faxed to Barclay 's requesting an appeal. Since this incident, I did speak to a manager named XXXX at Barclay 's. I faxed and sent the information certified backing my claim and it was denied a second time. All the letters are enclosed. When I tried speak with someone who actually investigates these cases, I got absolutely no response. Now the case has been reopened in Merchant Dispute at Barclay 's and I feel like my issue is falling on deaf ears. I also spoke to a Customer Service Manager at XXXX, his email is attached. All my documentation is attached. Even when I spoke to several people on Barclays end they confirmed they received my information I submitted based on findings but can not explain why the transaction wasn't reversed. It is very frustrating and I do not want my credit rating jeopardized by this outcome.My initial appeal letter is also attached. I have already spoken with a lawyer but wanted to try to resolve this initially with Barclays first. All documentation will be attached and scanned. A copy of my initial appeals letter is attached. Also attached are 6 letters received by Barclays. Two of the letters are conflicting, one saying they agree it was fraud ( dated XX/XX/XXXX ) the other dated XX/XX/XXXX upheld they're decision. A police report is attached, the detective agreed when he saw the tracking information that it was suspicious.} 
  & Prepaid card/debit card 
  & \parbox{5cm}{Prepaid card/debit card, Prepaid card/debit card, Prepaid card/debit card, Unknown, Prepaid card/debit card, Prepaid card/debit card, Debt collection, Prepaid card/debit card, Prepaid card/debit card, Prepaid card/debit card} \\
  \bottomrule
\end{tabular}
}
\label{tab_cc_annotations}
\end{table*}

\begin{table*}[htbp]
\caption{Sample annotations and ground truth values from the Wikipedia Movie Plots dataset. Note: Here \textbf{Unknown} means the annotator did not provide a label for that sample.}
\centering
\small
\resizebox{\textwidth}{!}{
\renewcommand{\arraystretch}{1.3}  % Optional: adds spacing between rows
\begin{tabular}{p{20cm}p{2cm}p{5cm}}
  \toprule
  \textbf{Text} & \textbf{Ground Truth} & \textbf{Annotations} \\
  \midrule

  \parbox{20cm}{Spoiled playboy Bob Merrick's (Rock Hudson) reckless behavior causes him to lose control of his speedboat. Rescuers send for the nearest resuscitator, located in Dr. Phillips's house across the lake. While the resuscitator is being used to save Merrick, Dr. Phillips suffers a heart attack and dies. Merrick ends up a patient at Dr. Phillips's clinic, where most of the doctors and nurses resent the fact that Merrick inadvertently caused Dr. Phillips's death. Helen Phillips (Jane Wyman), Dr. Phillips's widow, receives a flood of calls, letters, and visitors all offering to pay back loans that Dr. Phillips refused to accept repayment of during his life. Many claimed he refused by saying ""it was already used up."" Edward Randolph (Otto Kruger), a famous artist and Dr. Phillips's close friend, explains to Helen what that phrase means. This helps her to understand why her husband left little money, even though he had a very successful practice. Merrick discovers why everyone dislikes him. He runs from the clinic but collapses in front of Helen's car and ends up back at the hospital, where she learns his true identity. After his discharge, Merrick leaves a party, drunk. Merrick runs off the road and ends up at the home of Edward Randolph, who recognizes him. Randolph explains the secret belief that powered his own art and Dr. Phillips's success. Merrick decides to try out this new philosophy. His first attempt causes Helen to step into the path of a car while trying to run away from Merrick's advances. She is blinded by this accident. Merrick soberly commits to becoming a doctor, trying to fulfill Dr. Phillips's legacy. He also has fallen in love with Helen and secretly helps her adjust to her blindness under the guise of being simply a poor medical student, Robby. Merrick secretly arranges for Helen to travel to Europe and consult the best eye surgeons in the world. After extensive tests, these surgeons tell Helen there is no hope for recovery. Right after this, Robby shows up at her hotel to provide emotional support but eventually discovers that Helen has already guessed his real identity. Merrick asks Helen to marry him. Later that night, Helen realizes she will be a burden to him, and so runs away and disappears. Many years pass and Merrick is now a dedicated and successful brain surgeon who secretly continues his philanthropic acts, and searches for Helen. One evening, Randolph arrives with news that Helen is very sick, possibly dying, in a small Southwest hospital. They leave immediately for the hospital. Merrick arrives to find that Helen needs complex brain surgery to save her life. As the only capable surgeon at the hospital, Merrick performs this operation. After a long night waiting for the results, Helen awakens and discovers she can now see.} 
  & Drama
  & Drama, Drama, Drama, Drama, Drama, Drama, Drama, Unknown, Horror \\ \\

  \parbox{20cm}{Sports writer Steve Taggart (O'Neal) volunteers to do a series of articles for the Los Angeles Herald Examiner about a compulsive gambler he calls ""Mr. Green""...who is, in fact, himself. Taggart becomes obsessed with gambling in Las Vegas, which lands him deeply into debt. He compounds his money and gambling problems by dealing with associated loan sharks, including the mean and dangerous L.A. bookmaker known as ""The Dutchman"" (Chad Everett). Taggart soon learns that a local pro football quarterback, whose team he covers, is also on the Dutchman's payroll - as a means of cutting his own gambling debts. After clearing the story with his sports editor, John Saxon, Taggart journeys to Las Vegas for a field report on his gambling series; through a casino owner he meets a sexy casino cocktail hostess named Flo (Catherine Hicks). Loving the tables, at Flo's urging he gambles with Flo at roulette and wins, instead of taking her to bed. In Las Vegas Taggart also checks out assorted Las Vegas bookmakers, including Leroy. Taggart meets various Vegas gambling and business figures, including famed Las Vegas Sun publisher Hank Greenspun, for more insights into the Las Vegas gambling world. He is unaware that the Dutchman's tough guy enforcer, ""Panama Hat"" is closely following him until the Hat William Smith confronts him at the hotel pool as Taggart attempts to relax on a chaise lounge. Panama Hat orders Taggart to return to Los Angeles immediately, and settle up with the Dutchman, or there will be dire consequences. Taggart's risk-addiction and perennial gambling money-losses ultimately spill over into his personal life. After a day trip to Knott's Berry Farm Taggart brings his young daughter (Bridgette Andersen) to Hollywood Park; At the track pressbox, they see and chat with his colleagues, famed Los Angeles newspaper sportswriters Jim Murray and Alan Malamud. Taggart is now trying to stop gambling-but at the racetrack watching the races, he is physically assaulted by a track-goer to whom he owes money. Reporting to work the next day at the Herald Examiner, his newspaper editor (John Saxon) says he loves the ""Mr. Green"" series which the newspaper has been running, enough so that Saxon advances Taggart \$10,000 dollars for ""Mr. Green"" to use as seed money for more gambling. Upon more reflection on how truly dangerous gambling can be, Taggart visits Gamblers Anonymous in order to end his gambling compulsion. Taggart still returns to Las Vegas, where he becomes increasingly acquainted with Las Vegas high-roller Charley Peru (Giancarlo Giannini),in hopes of making a large score and breaking even. He also hopes of Peru helping him get the Dutchman's head thug, Panama Hat (William Smith), off Taggart's back. Finally Taggart decides to stop gambling ""forever"" Returning to Los Angeles, to celebrate ""kicking"" his gambling habit, Taggart places a few dollars into a slot machine at the Las Vegas Airport, where he magically scores a huge jackpot. Taggart immediately gets an attorney to hold the huge score in trust fund for his daughter. When he asks the attorney to reassure him ""even I cannot touch the money?"", his attorney replies, ""especially not you.""} 
  & Drama 
  & Drama, Unknown, Drama, Drama, Action, Drama, Horror, Comedy, Horror \\ \\

  \parbox{20cm}{One night, Caleb Colton (Adrian Pasdar), a young man in a small town, meets an attractive young drifter named Mae (Jenny Wright). Just before sunrise, she bites him on the neck and runs off. The rising sun causes Caleb's flesh to smoke and burn. Mae arrives with a group of roaming vampires in an RV and takes him away. The most psychotic of the vampires, Severen (Bill Paxton), wants to kill Caleb but Mae reveals that she has already turned him. Their charismatic leader Jesse Hooker (Lance Henriksen) reluctantly agrees to allow Caleb to remain with them for a week, to see if he can learn to hunt and gain the group's trust. Caleb is unwilling to kill to feed, which alienates him from the others. To protect him, Mae kills for him and then has him drink from her wrist. Jesse's group enters a bar and kills the occupants. They set the bar on fire and flee the scene. After Caleb endangers himself to help them escape their motel room during a daylight police raid, Jesse and the others are temporarily mollified, with Caleb asking Jesse how old he was and told he fought for the South. Caleb's father (Tim Thomerson) searches for Jesse's group. A child vampire in the group, Homer (Joshua John Miller) meets Caleb's sister Sarah (Marcie Leeds) and wants to turn her into his companion, but Caleb objects. While the group argues, Caleb's father arrives and holds them at gunpoint, demanding that Sarah be released. Jesse taunts him into shooting but regurgitates the bullet before wrestling the gun away. In the confusion, Sarah opens a door, letting in the sunlight and forcing the vampires back. Burning, Caleb escapes with his family. Caleb suggests they try giving him a blood transfusion to attempt to cure him. The transfusion successfully reverses Caleb's transformation. That night, the vampires search for Caleb and Sarah. Mae distracts Caleb by trying to persuade him to return to her while the others kidnap his sister. Caleb discovers the kidnapping and his tires slashed but gives chase on horseback. When the horse shies and throws him, he is confronted by Severen. Caleb commandeers a tractor-trailer and runs Severen over. The injured vampire suddenly appears on the hood of the truck and manages to rip apart the wiring in the engine. Caleb jackknifes the vehicle and jumps out as the truck explodes, killing Severen. Seeking revenge, Jesse and his girlfriend Diamondback (Jenette Goldstein) pursue him but are forced to flee in their car as dawn breaks. Not wanting Sarah to become another childlike monster, Mae breaks out of the back of the car with Sarah. Mae's flesh begins to smoke as she is burned by the sun but she carries Sarah into Caleb's arms, taking refuge under his jacket. Homer attempts to follow, but as he runs he dies from exposure to the sun. Jesse and Diamondback, their sunproofing ruined, also begin to burn. They attempt to run Caleb and Sarah over but fail, dying as the car blows up. Mae awakens later, her burns now healed. She too has been given a transfusion and is cured. She and Caleb comfort each other in a reassuring hug as the film ends.}
  & Horror 
  & Horror, Horror, Horror, Horror, Action, Horror, Drama, Drama, Comedy \\ \\

  \parbox{20cm}{Six friends, Christian, David, Kate, Johnny, Sara, and Melody, are traveling in an R.V. to get to the wedding of their friend, Kelly, in Galveston, Texas. However they become lost in a small town called Lovelock, and decide to spend the night at the local bed and breakfast, owned by the creepy Mr. Wise. While staying, the group insult the chef, Henri, causing an argument to break out. After everyone goes to bed, David goes to the kitchen to get a snack, only to discover Henri brutally murdered, before Mr Wise suffers a heart attack. With the phone line broken it takes until morning for the Sheriff, and his Deputy, Enus, to be summoned for help. The Sheriff is quick to suspect the group, and takes the keys to the R.V. away, so they can not leave the town until the investigation is over. The group goes into town, while the Sheriff arrests a mysterious drifter, who quickly becomes the prime suspect. The drifter warns Christian and Sara of ancient exotic wooden box, that Sara realizes belongs to Mr Wise. However it is too late, as Johnny arrives back at the bed \& breakfast and opens the box, unleashing the ""Kuman Thong"" which possesses him, causing him to savagely murder various people. Meanwhile, Christian and Sara meet town local Lisa Belmont (Miranda Bailey) who swears she saw Mr Wise dig up the body of his dead son and perform a form of black magic on the body. Sara and Christian return to the bed \& breakfast and discover Johnny has opened the box. Sara and Christian alert the Sheriff, who drives them to a local party, that David, Kate and Melody are attending with the rest of the town folk. The possessed Johnny arrives and a bloody massacre ensues, with the town folk, including Enus, becoming zombies as Johnny puts various body parts of his victims in the box. In the chaos, Christian is decapitated as David, Kate, Sara and Melody escape with the Sheriff in a truck. They accidentally run over the drifter, knocking him unconscious after he escapes from his prison cell. Taking him with them, the radiator soon blows in the truck, forcing the group to take shelter in the bed \& breakfast. They gather weapons, before the drifter tells them they must retrieve the body of Mr Wise to kill Johnny. The zombies arrive at the bed \& breakfast and the group fend them off, before the Sheriff, Melody and the drifter sneak out the back door to retrieve the bones of Mr Wise. At the bed \& breakfast, the zombies retrieve some of David's blood that was on the step of the house, placing it in the box causing him to become possessed. David beats Kate to death with a metal pole, before attacking Sara. However Sara manages to kill David with a chainsaw. At the cemetery, the Sheriff, Melody and the drifter retrieve the body of Mr Wise and perform a black magic spell, taking the bones from the body. As they travel to the bed \& breakfast they encounter a group of zombies. The Sheriff has his neck snapped, killing him, before Lisa arrives and rescues the drifter and Melody, who continue on to the bed \& breakfast. Meanwhile, at the bed \& breakfast, the zombies break in. Sara fights them, but is soon cornered. The drifter, and Melody arrive outside, where Melody shoots Johnny through the heart with a bone from the body of Mr Wise, killing him and the other zombies. Sara reunites with Melody and the drifter, and together they leave Lovelock in their R.V.} 
  & Comedy 
  & Horror, Drama, Drama, Horror, Horror, Comedy, Comedy, Drama, Action \\ \\

  \parbox{20cm}{The film revolves around Kathavarayan (Karan), who sells illicit arrack in Hogenakkal. Though he is involved in arrack trade, he is a man with golden heart. Malathy (Vidisha), a student, comes to the village as part of her NSS project. She vows to end the arrack menace in the village and takes efforts to arrest Kathavarayan. Thanks to her efforts, Kathavarayan gets arrested and lodged in prison in Chennai. Call it fate, Malathy is arrested for no fault of her in Chennai. Her efforts to expose drug-pedlers in Chennai backfires. They hatch a conspiracy and Malathy falls a prey to it. She gets arrested for possessing drugs and lands in prison. Kathavarayan comes out of jail to avenge Malathy. But comes to know her real intentions and the trouble she is facing. He masterminds a plan and joins the gang selling drugs and eventually exposes them only to save Malathy.} 
  & Action 
  & Drama, Action, Action, Action, Action, Drama, Unknown, Action, Drama \\

  \bottomrule
\end{tabular}
}
\label{tab_wiki_annotations}
\end{table*}

\subsection{Full Data Collection Protocol}
\label{subsec_protocol}

We provide the full data collection protocol (for the Consumer Complaint dataset) in this section (preserving author and institution anonymity). The protocols for the other datasets are similar. \\

\begin{center}
    \textbf{Instructions for Data Annotation}
\end{center}

\noindent You are being asked to voluntarily participate in a research study, funded by \textbf{XX}. Researchers at \textbf{XX} are studying the development of artificial intelligence (AI) models under real-world challenges such as crowdsourced data annotations. In this experiment, you will be shown $3,000$ short text snippets. Each snippet represents a complaint filed by a customer and can belong to one of the following $6$ categories:\\

\begin{itemize}

\item Debt collection (1) 
\item Prepaid card/debit card (2)
\item Mortgage (3)
\item Checking/savings account (4)
\item Student loan (5)
\item Vehicle loan/lease (6)

\end{itemize}

\noindent \newline Your task will be to annotate each text sample with its most appropriate category (based on the topic of the complaint) from the above list. You can also abstain from labeling, if you are not sure about the category. Use the number in the parentheses next to each category, to annotate each sample. If you want to abstain from labeling a particular sample, please use $0$. \\

\noindent Please note the following: 

\begin{itemize} 

\item Your involvement in the study is expected to take about $20 - 25$ hours of your time
\item You are free to do it over multiple sessions. However, you are required to annotate the samples sequentially (starting from sample $1$ through sample $3,000$)
\item You will receive $4$ cents ($\bold{0.04}$ USD) for annotating each sample. However, you will receive your complete payment of $120$ USD only after you complete annotating all the $3,000$ samples; if you discontinue taking part in the study midway, you will not receive any payment
\item Please annotate each sample with exactly one category (or abstain of you are not sure)

\end{itemize}

\noindent \newline We will not record your name or any information that shows your identity. If you have any questions, please contact: \textbf{XX}

\subsection{Performance Analysis with the GPT-2 Backbone}
\label{subsec_GPT2}

\begin{figure}[ht]
	\centering
         % \label{fig_agnews_GPT}
          \includegraphics[trim = 1.3in 3.2in 1.2in 3.4in,clip,width=.4\textwidth]{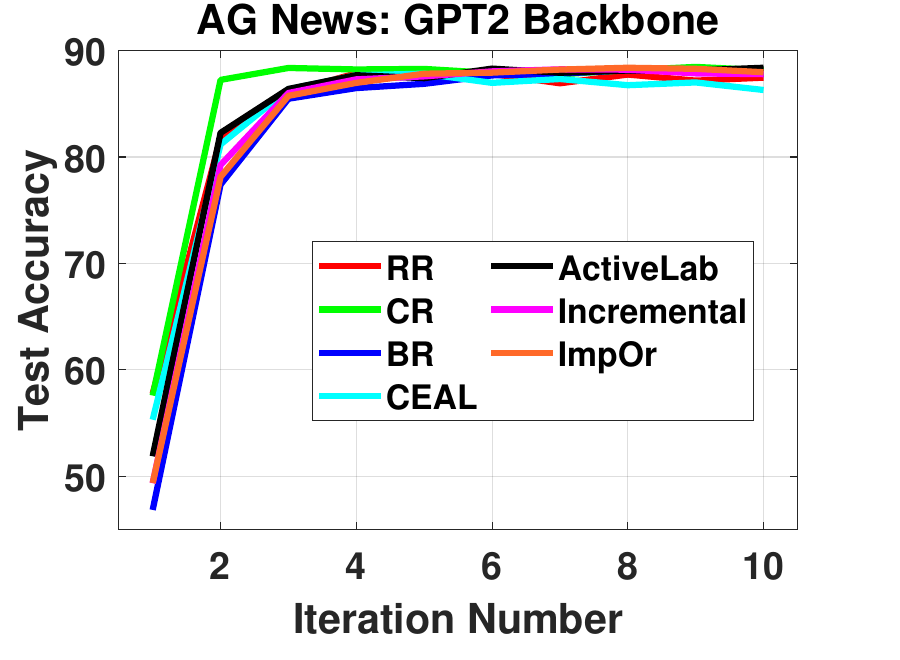}   
          \caption{Performance of AL algorithms using the GPT-2 Backbone on the AG News dataset. Best viewed in color.}
        \label{fig_GPT2_results}     
\end{figure}

In this section, we analyze the performance of the active learning algorithms using the GPT-2 backbone \cite{GPT2_Paper}. The results are presented in Figure \ref{fig_GPT2_results}. The \textit{ActiveLab} and \textit{CR} methods depict the best performance for this experiment, although the difference in accuracy with the other methods is lower, compared to Figure \ref{fig_agnews}.

%\end{comment}

\end{document}